\documentclass{bmvc2k}

\title{Continuous-time 4D Reconstruction via Event-guided Latent Feature Interpolation}

\addauthor{Gwangtak Bae}{tak3452@snu.ac.kr}{1}
\addauthor{Seunggu Kang}{jrkangsg@snu.ac.kr}{2}
\addauthor{Junho Kim}{82magnolia@snu.ac.kr}{1}
\addauthor{Eun Sun Lee}{eunsunlee@snu.ac.kr}{1}
\addauthor{Changwoon Choi}{changwoon.choi00@gmail.com}{1}
\addauthor{Young Min Kim$^\dagger$}{youngmin.kim@snu.ac.kr}{1,3}

\addinstitution{
 ECE\\
 Seoul National University, South Korea
}
\addinstitution{
 FAM\\
 Seoul National University, South Korea
}
\addinstitution{
 INMC \& IPAI\\
 Seoul National University, South Korea
}

\footnotetext{Young Min Kim is the corresponding author.}

\runninghead{BAE et al.}{Continuous-time 4D Recon. via Event-guided Feature Interp.}

\usepackage[capitalize]{cleveref}
\usepackage{amsmath}
\usepackage{amssymb}
\usepackage{bbm}
\usepackage{multirow}
\usepackage{booktabs}
\usepackage[export]{adjustbox}
\usepackage{graphicx}

\begin{document}

\maketitle

\begin{abstract}
Recent geometric foundation models can successfully convert a video sequence into frames of registered point clouds, but cannot predict moments beyond the observed temporal resolution.
We propose a continuous-time 4D reconstruction method that extracts 3D structure and camera poses at any time instant, with additional guidance from event-camera measurements.
While event cameras provide evidence at microsecond resolution to continuously fill the temporal gap,  integrating them into existing RGB-based foundation models remains challenging.
We formulate the problem as interpolation in the latent feature space and augment event-based losses in the image space as an auxiliary task to boost accuracy.
Operating in the latent space leverages the powerful geometric priors of foundation models, while the event loss provides physically grounded guidance.
To effectively address the lack of paired training data, we further propose a self-supervised training framework that uses pseudo-ground-truth labels of estimated geometry and simulated events generated from high-frame-rate RGB videos.
Through extensive experiments, we demonstrate that the proposed pipeline outperforms baselines that temporally interpolate either the input RGB videos or the estimated geometry, and that it generalizes to unseen real-world scenarios.
Controlled experiments with synthetic corruptions show that it remains robust under motion-blur and low-light conditions.
\end{abstract}

\begin{figure*}[t]
    \centering
    \includegraphics[width=1.0\linewidth]{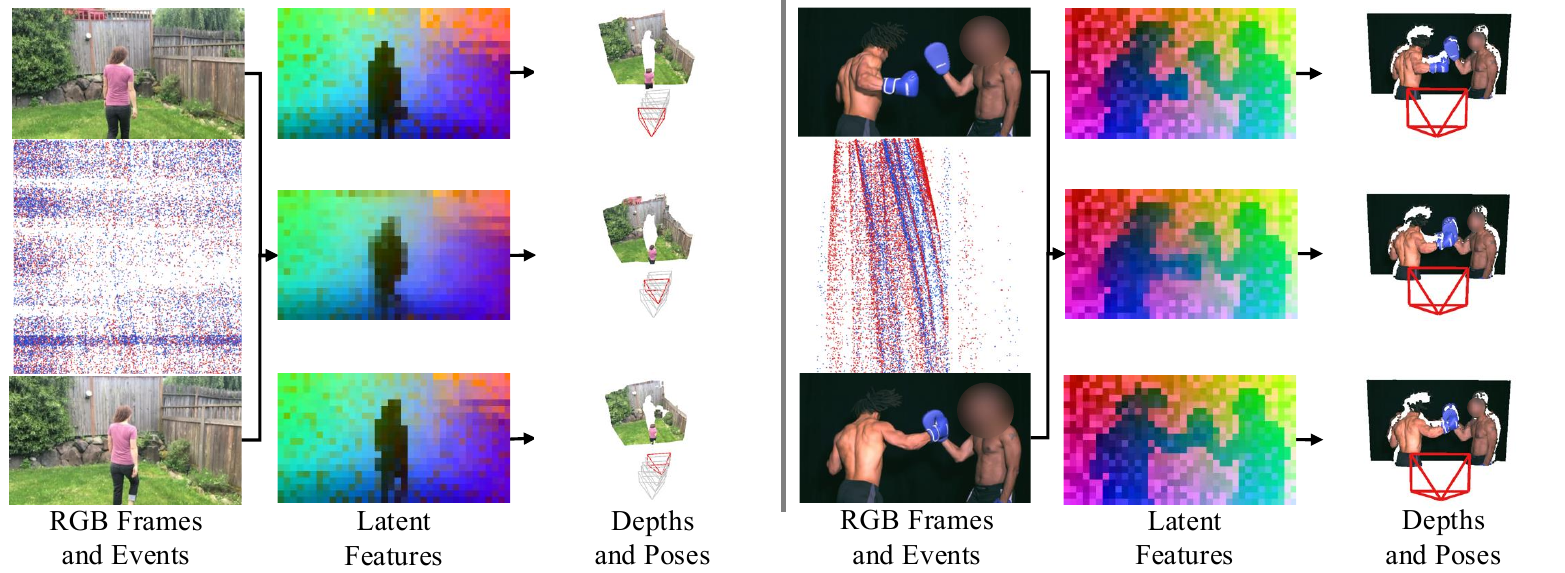}
    \vspace{-1.5em}
    \caption{We propose an event-guided latent feature interpolation pipeline for continuous-time 4D reconstruction.
    Our method uses the event stream to interpolate the latent feature of a pretrained RGB-based geometric foundation model, which carries rich geometric information.
    The interpolated feature is decoded into depth and camera pose at a given timestamp.
    }
    \vspace{-0.5em}
    \label{fig:teaser}
\end{figure*}

\section{Introduction}

Recent geometric foundation models~\cite{wang2024dust3r, wang2025vggt, lin2025depth, cong2025e3d} demonstrate the ability to reconstruct geometry from RGB images and videos.
Trained on large-scale datasets, the strong geometric priors successfully estimate frame-wise depth and camera pose across diverse scenes.
Such dynamic structural knowledge, combined with rich semantic context, enables image observation to serve as a sufficient input modality for practical downstream applications, including robotics~\cite{allshire2025visual}, AR/VR~\cite{jiang2024vr}, and embodied AI~\cite{zhen2025learning}.
However, because their geometric predictions are tied to the input frames, they cannot capture fast, subtle dynamics beyond the original temporal resolution.

We propose using a stream of RGB video, paired with event-camera measurements, for continuous-time 4D reconstruction, enabling recovery of 3D geometry and camera motion at arbitrary timestamps.
Event cameras asynchronously record local brightness changes with microsecond-level temporal resolution~\cite{gallego2020event}, providing precise evidence of visual changes between frames.
However, paired event-RGB data with ground-truth geometry is scarce to train a foundation model from scratch.
A plausible solution is to integrate asynchronous event measurements into the strong geometric priors of RGB-based models.

We frame the continuous-time 4D reconstruction problem as temporal interpolation: reconstructing the geometry and camera motion at an interpolated timestamp, given two adjacent RGB frames and the event stream between them.
If we aim to incorporate event input without altering the foundation model, we can either interpolate the input image or the 3D output.
\emph{Input-space interpolation} synthesizes intermediate RGB frames using event-based video frame interpolation~\cite{tulyakov2021time, tulyakov2022time, chen2025repurposing} and feeds them into a frame-based geometric foundation model.
Because the interpolation is optimized for image quality, the results exhibit geometric artifacts around occlusions, motion boundaries, and blurred regions, leading to errors in depth and pose.
At the other end, \emph{output-space interpolation}~\cite{guo2026interp3r} interpolates the reconstructed pointmaps using event measurements.
However, the decoded output is agnostic to rich internal representations, which are useful for modeling subtle inter-frame dynamics, such as multi-view correspondence and contextual scene cues.

In contrast, we propose \emph{latent-space interpolation}, which performs continuous-time interpolation in the internal latent feature space of the geometric foundation model.
The conventional model operates frame-wise: the encoder converts each frame into a latent representation, and the decoder extracts geometric information, such as point maps, depth, and camera pose, for that frame.
Given two adjacent RGB frames and an event stream between them, our framework converts the frame-wise features into a continuous feature field volume that can be sampled at any timestamp and decoded into geometry, illustrated in Figure~\ref{fig:teaser}.
The latent interpolation further incorporates event evidence to match temporal information in image space.
In parallel with our 4D reconstruction pipeline, we attach a color interpolation branch that predicts continuous-time color information at arbitrary timestamps, whose appearance change should match the measurements in event streams.
The event formulation naturally handles images with challenging illumination or fast motion, given the sensor's measurement characteristics.
Most importantly, the appearance constraint serves as a proxy task, providing useful temporal constraints that guide underlying geometric changes.

While we build on existing foundation models, we still need to train the proposed framework using high-frequency ground-truth depth and pose information acquired from an RGB video paired with an event camera, which is rarely available.
We circumvent the problem by using pseudo-ground-truth labels obtained from high-frame-rate videos.
Temporally dense RGB frames can simulate realistic events and provide estimated depth and camera pose using geometric foundation models.
Empirically, the trained model successfully operates on both real-world and synthetic datasets, yielding more accurate 4D reconstruction than both input-space and output-space interpolation baselines.
We further analyze its robustness to motion blur and low light, a direct consequence of incorporating event measurements.

In summary, our main contributions are as follows:
\begin{itemize}
    \item We propose event-guided latent feature interpolation, a framework for continuous-time 4D reconstruction that interpolates within the feature space of a pretrained geometric foundation model, fully leveraging its learned geometric prior to decode depth and camera pose at arbitrary timestamps.
    \item We introduce a proxy task of continuous appearance interpolation and event simulation, which simulates events from the predicted continuous color changes and matches them to the observed event stream, regularizing geometry interpolation with an additional temporal constraint.
    \item We propose a self-supervised training framework built on high-frame-rate RGB datasets that simulates realistic events and distills pseudo-ground-truth depth and pose from a pretrained foundation model, addressing the scarcity of event datasets with ground-truth geometry while transferring effectively to real-world event streams.
    \item We demonstrate through experiments on real-world datasets and synthetic motion-blur and low-light corruptions that our latent-space interpolation yields more accurate 4D reconstruction than RGB-interpolation and pointmap-interpolation baselines across diverse scenarios.
\end{itemize}

\section{Related Works}

\paragraph{3D/4D Geometric Foundation Model.}
Beyond the classical SfM/MVS pipeline, recent geometric foundation models learn to predict scene geometry directly from images via large-scale training.
DUSt3R~\cite{wang2024dust3r} regresses per-pixel 3D points from an image pair in a shared coordinate frame, implicitly recovering relative camera poses and showing strong generalization across diverse scenes.
Building on this representation, MASt3R~\cite{leroy2024grounding} adds dense feature matching for accurate two-view correspondence, VGGT~\cite{wang2025vggt} scales to many views with a feed-forward transformer that jointly predicts depth, pose, and tracks, and MapAnything~\cite{keetha2025mapanything} unifies varied inputs such as single images, multi-view collections, and posed sequences within a single framework.
Extending pointmap-based models from static to dynamic scenes, MonST3R~\cite{zhang2025monst3r} finetunes DUSt3R on dynamic datasets and CUT3R~\cite{wang2025continuous} processes streaming video with a recurrent persistent state.
Depth Anything 3 (DA3)~\cite{lin2025depth}, which we adopt as a backbone, jointly predicts depth and camera rays under a unified depth-ray representation on a plain transformer, achieving state-of-the-art results on both static 3D and dynamic 4D reconstruction.
Beyond per-frame reconstruction, recent works directly output continuous-time 4D representations~\cite{liang2024feed,ma20264d,zhang2025efficiently}, but rely solely on sparse RGB frames without any direct measurement within the blind interval between them.
A parallel line of work injects event streams into geometric foundation models~\cite{peng2026adae,wu2026eag3r}, yet they treat events as a per-frame robustness cue and do not exploit their microsecond-scale temporal resolution for continuous-time reconstruction.

\paragraph{Event-guided Continuous-time Interpolation.}
Since event cameras provide direct observations within the unknown interval between RGB frames, many works have leveraged them for continuous-time interpolation.
Event-based video frame interpolation (EVFI) treats the event stream as motion guidance to synthesize intermediate RGB frames between two frames.
TimeLens~\cite{tulyakov2021time} pioneered this direction by combining event-driven motion estimation with frame-based warping in a unified network.
TimeLens++~\cite{tulyakov2022time} extends it with parametric motion trajectories, and more recently, VDM-EVFI~\cite{chen2025repurposing} adapts pretrained video diffusion models to event-conditioned interpolation.
However, feeding such synthesized intermediate frames directly into a frame-based geometric model amplifies their visual artifacts around occlusions and motion boundaries, leading to downstream depth and pose errors.
Closest to ours, Interp3R~\cite{guo2026interp3r} instead interpolates 3D geometry rather than color, using events to predict pointmaps at arbitrary times between two frames.
However, Interp3R operates on the model's already-decoded pointmaps, after the rich latent representation has been collapsed into a 3D output, discarding intermediate cues such as multi-view correspondence and contextual scene information.
In contrast, we interpolate in the latent feature space between the encoder and decoder of a pretrained geometric foundation model, preserving its learned geometric prior while allowing events to guide fine-grained temporal changes.

\section{Method}
\label{sec:method}

\subsection{Overview}
\label{subsec:overview}

Given a set of RGB frames $\{I_0,\dots,I_N\}$ and the event stream $E$ recorded across them, our goal is to estimate depth and camera pose at an arbitrary query timestamp $t$ within any inter-frame interval.
We build on Depth Anything 3 (DA3)~\cite{lin2025depth}, an RGB-based feed-forward geometric foundation model, which we use as a frozen reconstruction backbone.
Its transformer encoder jointly maps all input frames to per-frame latent feature tokens, $\{F_0,\dots,F_N\} = \mathrm{Enc}(I_0,\dots,I_N)$, which carry rich geometric and contextual information.
A DPT~\cite{ranftl2021vision} decoder then maps each latent feature to a depth map and a ray map, $(D_i, R_i) = \mathrm{Dec}(F_i)$.
The camera pose is recovered from the per-pixel rays $R_i$, while the depth $D_i$ together with the rays yields a per-frame point map $X_i$.

We perform temporal interpolation directly in the latent token space between the encoder and decoder, where rich geometric information is still preserved.
To recover the geometry within each blind interval, we perform interpolation independently for each adjacent frame pair.
Given two adjacent RGB frames $(I_s, I_e)$, their features $(F_s, F_e)$, and the event stream $E$ between them, our network builds a continuous latent feature field $F(t)$ over the inter-frame interval, with the endpoint features $F_s$ and $F_e$ serving as boundary anchors at $t=0$ and $t=1$.
At a query timestamp $t$, we sample $F(t)$ and decode it with the frozen decoder to obtain depth and camera pose, $(D_t, R_t) = \mathrm{Dec}(F(t))$.

\begin{figure*}[t]
    \centering
    \includegraphics[width=1.0\linewidth]{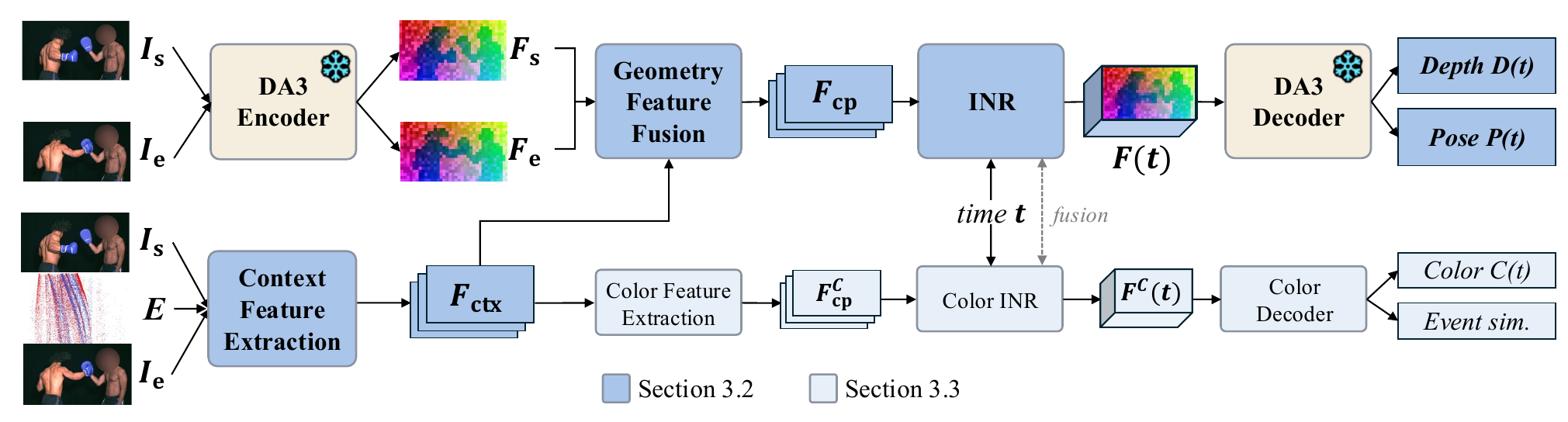}
    \vspace{-1.5em}
    \caption{Network architecture. From the two adjacent frames $I_s, I_e$ and the events $E$ between them, we first extract a temporal-context feature $F_{\mathrm{ctx}}$ that encodes the inter-frame temporal change.
    This context is then fused with the endpoint features $F_s, F_e$, producing control-point features $F_{\mathrm{cp}}$, which an INR block samples and refines into the interpolated feature $F(t)$ at the query timestamp $t$.
    An appearance interpolation branch is jointly trained as a proxy task to further regularize the latent feature trajectory.
    }
    \vspace{-1.5em}
    \label{fig:network}
\end{figure*}

As illustrated in~\cref{fig:network}, we perform interpolation in the pretrained feature space while keeping the DA3 encoder and decoder frozen.
Given the input RGB frames and events $I_s$, $I_e$, and $E$, the Context Feature Extraction module produces a temporal-context feature $F_{\mathrm{ctx}}$.
The Geometry Feature Fusion module then fuses $F_{\mathrm{ctx}}$ with the endpoint DA3 features $F_s$ and $F_e$ to obtain the control-point features $F_{\mathrm{cp}}$.
Through implicit neural representation (INR)-based sampling and refinement, we obtain the feature $F(t)$ at a query timestamp $t$, which is decoded into depth and camera pose.
The color interpolation modules, shown in skyblue in~\cref{fig:network}, are jointly trained to regularize the geometry feature interpolation.

\cref{subsec:network} details the event-guided feature interpolation network, which extracts temporal guidance from the event stream and RGB inputs to produce the continuous latent feature field $F(t)$.
\cref{subsec:proxy} introduces an appearance-based proxy task that jointly predicts continuous color and simulates events based on it, thereby further regularizing the latent trajectory.
In~\cref{subsec:training}, we explain our self-supervised framework that distills pseudo-ground-truth geometry from the frozen DA3 teacher and simulates events from high-frame-rate RGB video, requiring no real event or geometry annotations.

\subsection{Event-Guided Feature Interpolation Network}
\label{subsec:network}

Our network takes $\{I_s, I_e, F_s, F_e, E\}$ as input and predicts the interpolated feature $F(t)$ at a query time $t$, which is then converted into depth and camera pose by the frozen decoder.
Here $I_s, I_e \in \mathbb{R}^{3\times H\times W}$ are the RGB endpoints, $E$ is the event stream over the interval, and $F_s, F_e$ are the corresponding DA3 features $F\in\mathbb{R}^{S\times H_f \times W_f \times C}$, where $S$ is the number of scales, $H_f$ and $W_f$ are the height and width of the feature grid, and $C$ is the channel dimension.
The $S=4$ scales are the multi-level feature maps extracted from different ViT~\cite{dosovitskiy2020image} layers and consumed by the frozen DPT~\cite{ranftl2021vision} decoder.
Since each feature map may carry different semantic and geometric information, we train four separate Geometry Feature Fusion and INR modules in~\cref{fig:network}, while sharing a single Context Feature Extraction module across scales.
In the following, we describe the processing of a single scale for clarity.
As illustrated in~\cref{fig:network}, the network has three major modules: (i) the Context Feature Extraction module turns the two frames and the events between them into a temporal-context feature $F_\mathrm{ctx}$; (ii) the Geometry Feature Fusion module aggregates the endpoint features $F_s, F_e$ under this context into $K$ discrete control-point features $F_\mathrm{cp}$; and (iii) the INR (Implicit Neural Representation) module samples and refines them into the continuous feature $F(t)$.

\paragraph{Context feature extraction.}
We fuse the two adjacent frames and the events between them into a temporal context feature that carries the inter-frame motion.
The key design is to treat the $K$-bin event voxel as a $K$-frame video, so that the temporal change within the interval is processed as a sequence.
Concretely, given events $\{(x_k, y_k, t_k, p_k)\}$ over the interval $[t_s, t_e]$ with polarities $p_k\in\{-1,+1\}$, we accumulate them into the voxel grid with bilinear interpolation in time as in~\cite{zihao2018unsupervised}.
Modality-specific stems separately encode the two RGB frames and each event bin. The RGB features are blended across the $K$ steps by time-based weights $w_i = i/(K-1)$, and the two modalities are then fused into a single context sequence $F_{\mathrm{ctx}}\in\mathbb{R}^{K\times H_f \times W_f \times C}$.
We then use a recurrent encoder from the video restoration literature, RVRT-style propagation~\cite{liang2022recurrent}, to sequentially aggregate temporal cues across the $K$ control points of $F_{\mathrm{ctx}}$.

\paragraph{Geometry feature fusion.}
Conditioned on this temporal context, we aggregate the relevant geometric information from the two endpoint features $F_s, F_e$.
Using the context feature $F_{\mathrm{ctx}}$ as the query and the endpoint features as keys and values, we cross-attend to $F_s$ and $F_e$ separately and blend the two retrievals with a content-aware gate $\alpha$ initialized to the temporal prior $t$.
Because the endpoints describe only the two boundary timestamps, we further add $F_{\mathrm{ctx}}$ back as a residual, supplying event-driven cues about the intermediate moments of the interval.
The two operations combine into the control-point features $F_{\mathrm{cp}}\in\mathbb{R}^{K\times H_f \times W_f \times C}$, one per control point,
\begin{equation}
\label{eq:fusion}
F_{\mathrm{cp}} = (1-\alpha)\odot \mathrm{CA}(F_{\mathrm{ctx}}, F_s) + \alpha\odot \mathrm{CA}(F_{\mathrm{ctx}}, F_e) + F_{\mathrm{ctx}},
\end{equation}
where the cross-attention block $\mathrm{CA}(\cdot,\cdot)$ takes the first argument as the query and the second as the key/value.

\paragraph{Continuous output representation.}
To obtain a continuous feature at any query time $t$, we perform INR-based refinement over the $K$ control-point features $F_{\mathrm{cp}}$.
We first sample the two control points adjacent to $t$ and blend them into the feature at $t$, then encode $t$ with a frequency embedding~\cite{mildenhall2021nerf} and refine the feature through FiLM-modulated~\cite{perez2018film} layers, where an MLP maps the time embedding to per-layer affine parameters.

For stable training, and inspired by zero-initialized residual learning~\cite{zhang2023adding}, we formulate the INR output as a residual over a linear-interpolation baseline of the endpoint features.
Specifically, the baseline is defined as $F_{\mathrm{Lerp}}(t) = (1-t)\,F_s + t\,F_e$, and the output projection of the INR is zero-initialized so that training starts exactly from this baseline:
\begin{equation}
\label{eq:inr}
F(t) = F_{\mathrm{INR}}(F_{\mathrm{cp}}, t) + F_{\mathrm{Lerp}}(t).
\end{equation}
Here, $F_{\mathrm{INR}}(F_{\mathrm{cp}}, t)$ samples $F_{\mathrm{cp}}$ at $t$ and refines it through the FiLM layers.
The frozen decoder maps $F(t)$ to a depth map and a ray map, $(D(t), R(t)) = \mathrm{Dec}(F(t))$, and the camera pose is recovered from the ray map.

\paragraph{Geometry losses.}
We supervise the interpolated depth and ray maps with pseudo-ground-truth labels.
For the depth loss, following MiDaS~\cite{ranftl2020towards}, we use a scale-and-shift-invariant loss in disparity space.
We align the prediction with a single affine transform $(s, b)$ fit per sequence, shared across all intermediate frames for temporal consistency, and define
\begin{equation}
\label{eq:depth}
L_{\mathrm{depth}} = \lVert s\,\hat{p} + b - p^* \rVert_{1} + \lambda_{\mathrm{grad}}\, L_{\mathrm{grad}},
\end{equation}
where $\hat{p}$ and $p^*$ are the predicted and pseudo-ground-truth disparities, large-residual pixels are trimmed, and $L_{\mathrm{grad}}$ is a multi-scale gradient term in~\cite{ranftl2020towards}.
For the ray loss, we apply a $\mathrm{Sim}(3)$-invariant alignment between the predicted ray origins and directions $(\hat{\mathbf{o}}, \hat{\mathbf{d}})$ and the ground-truth $(\mathbf{o}, \mathbf{d})$.
We estimate the rotation $\mathbf{R}$ from the ray directions, then the scale $s$ and translation $\mathbf{t}$ from the ray origins, $(s, \mathbf{R}, \mathbf{t}) = \mathrm{Sim3Align}(\{\hat{\mathbf{o}}, \hat{\mathbf{d}}\}, \{\mathbf{o}, \mathbf{d}\})$, and apply it as
\begin{equation}
\label{eq:ray-align}
\tilde{\mathbf{o}} = s\,\mathbf{R}\,\hat{\mathbf{o}} + \mathbf{t}, \qquad \tilde{\mathbf{d}} = \mathbf{R}\,\hat{\mathbf{d}} .
\end{equation}
The loss combines a Charbonnier penalty $\rho$ on the origin residual with a direction term,
\begin{equation}
\label{eq:ray}
L_{\mathrm{ray}} = \lambda_{o}\rho\big(\tilde{\mathbf{o}}(\mathbf{u}) - \mathbf{o}(\mathbf{u})\big) + \lambda_d\big(1 - \langle \tilde{\mathbf{d}}(\mathbf{u}),\, \mathbf{d}(\mathbf{u})\rangle\big),
\end{equation}
where $\langle\cdot,\cdot\rangle$ is the inner product between unit ray directions.

\subsection{Appearance-based Proxy Task}
\label{subsec:proxy}

\paragraph{Joint appearance-geometry estimation.}
The appearance proxy is motivated by the intuition that temporal appearance changes carry cues about the underlying geometric changes, making continuous color prediction a natural regularizer of the latent feature trajectory.
We thus attach a continuous color interpolation branch alongside the geometry branch, predicting continuous-time color $C(t)$ at the same query timestamps as depth and camera pose.
The color branch mirrors the depth branch, sharing the context feature $F_{\mathrm{ctx}}$ but running its own Color Feature Extraction, Color INR and Color Decoder modules, as depicted in~\cref{fig:network}.
The two branches are coupled both through this shared context and through gated cross-modal injection between their INR heads.
At each INR layer, one branch adds the other's query-time feature through a learned gate, initialized near zero so the branches start decoupled and learn to couple over training.
We supervise $C(t)$ with a color reconstruction loss combining Charbonnier, MS-SSIM, and LPIPS terms,
\begin{equation}
\label{eq:color}
L_{\mathrm{color}} = L_{\mathrm{Charb}} + \lambda_{\mathrm{SSIM}}\,L_{\mathrm{MS\text{-}SSIM}} + \lambda_{\mathrm{LPIPS}}\,L_{\mathrm{LPIPS}} .
\end{equation}

\begin{figure*}[t]
    \centering
    \includegraphics[width=0.9\linewidth]{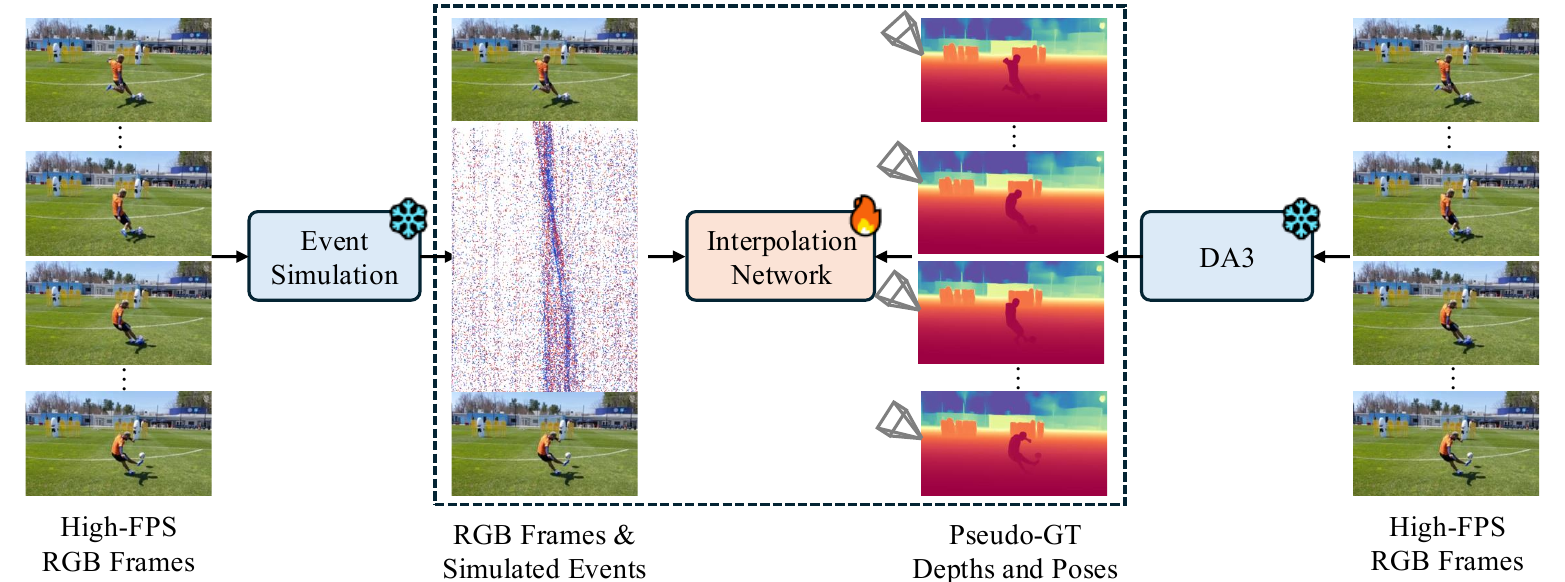}
    \vspace{-0.5em}
    \caption{Self-supervised training framework.
    From a high-frame-rate RGB video, we simulate realistic input events and distill pseudo-GT depth and camera pose from a pretrained geometric foundation model.
    Combining event simulation and self-distillation lets us construct large and diverse training data without any real-world event or geometry annotations.
    }
    \vspace{-0.5em}
    \label{fig:training}
\end{figure*}

\paragraph{Event simulation loss.}
We regularize the continuous color trajectory to be consistent with the observed events through an event simulation loss.
An event fires at a pixel $\mathbf{u}$ when its log-intensity change since the last event exceeds a contrast threshold $C_{\mathrm{thres}}$ that varies with the sensor and the sensing conditions, with polarity $p\in\{-1,+1\}$, giving microsecond-scale temporal cues.
Because our color branch is continuous in time, we can simulate this process at fine temporal resolution by querying it at two arbitrarily nearby timestamps $t$ and $t+\Delta t$ for the per-pixel intensities $I(\mathbf{u}, t)$ and $I(\mathbf{u}, t+\Delta t)$ and taking their log-intensity change,
\begin{equation}
\label{eq:dlog}
\Delta I(\mathbf{u}, t) = \log I(\mathbf{u}, t+\Delta t) - \log I(\mathbf{u}, t) .
\end{equation}
We then accumulate the polarities of the observed events at $\mathbf{u}$ over $[t, t+\Delta t]$ into a signed count,
\begin{equation}
\label{eq:esum}
E(\mathbf{u}) = \sum_{k\in\mathcal{K}(\mathbf{u})} p_k, \quad p_k\in\{-1,+1\} ,
\end{equation}
where $\mathcal{K}(\mathbf{u})$ is the set of events triggered at pixel $\mathbf{u}$ within $[t, t+\Delta t]$ and $p_k$ are their polarities.
Since the contrast threshold $C_{\mathrm{thres}}$ is not known precisely, instead of a single fixed value we use a tolerance band $[C_{\mathrm{low}}, C_{\mathrm{high}}]$, which gives the lower and upper bounds of the log-intensity change consistent with $E(\mathbf{u})$,
\begin{equation}
\label{eq:band}
T_{\min}(\mathbf{u}) = \min\!\big(E(\mathbf{u})C_{\mathrm{low}},\, E(\mathbf{u})C_{\mathrm{high}}\big), \qquad
T_{\max}(\mathbf{u}) = \max\!\big(E(\mathbf{u})C_{\mathrm{low}},\, E(\mathbf{u})C_{\mathrm{high}}\big) .
\end{equation}
The event simulation loss keeps $\Delta I$ within $[T_{\min}, T_{\max}]$ at pixels where events are observed, and below $C_{\mathrm{high}}$ at pixels where no event fires,
\begin{equation}
\label{eq:event}
\begin{split}
L_{\mathrm{event}} = {}& \mathbbm{1}[E(\mathbf{u})\neq 0]\Big(\max\big(0, T_{\min}(\mathbf{u}) - \Delta I\big) + \max\big(0, \Delta I - T_{\max}(\mathbf{u})\big)\Big) \\
& + \mathbbm{1}[E(\mathbf{u}) = 0]\,\max\big(0, \lvert\Delta I\rvert - C_{\mathrm{high}}\big) .
\end{split}
\end{equation}
Since this loss relies only on the input events, it applies at any timestamp $t$ sampled in $[0, 1]$, not just the intermediate frames where pseudo-GT geometry exists.

\paragraph{Total objective.}
The full objective combines the geometry losses with the appearance proxy,
\begin{equation}
\label{eq:total}
L = L_{\mathrm{depth}} + \lambda_{\mathrm{ray}} L_{\mathrm{ray}} + \lambda_{\mathrm{color}} L_{\mathrm{color}} + \lambda_{\mathrm{event}} L_{\mathrm{event}} .
\end{equation}

\subsection{Self-Supervised Training from High-Frame-Rate Video}
\label{subsec:training}

Unlike the RGB domain, building a large-scale training set for event-guided 4D reconstruction is hard, since real RGB-event datasets with ground-truth depth and camera pose are scarce and limited in scene diversity.
To obtain a large and diverse training set, we propose a self-supervised training pipeline based on high-frame-rate video data and self-distillation.
High-frame-rate RGB video is relatively abundant and already used to train color interpolation networks~\cite{su2017deep, chen2022videoinr, shen2020blurry, chen2023sportsslomo}.
From these dense frames we synthesize both the input events and the ground-truth geometry.
As shown in~\cref{fig:training}, we first simulate realistic events from the dense frames with an event simulator~\cite{gehrig2020video}.
We then obtain pseudo ground-truth depth and camera pose by distilling the frozen DA3 teacher, i.e., running it on the dense frames to get frame-wise depth and camera pose.
Pairing the simulated events with this pseudo-ground-truth lets us train across diverse scenes without any real-world event data.

\section{Experimental Results}
\label{sec:experiments}

\subsection{Experimental Setup}

\paragraph{Implementation details.}
Our network is built on top of Depth Anything 3 (DA3)~\cite{lin2025depth}, specifically the DA3-LARGE-1.1 model.
The number of temporal control points is set to $K = 20$, and $C_{\mathrm{low}}$ and $C_{\mathrm{high}}$ are set to 0.15 and 0.35.
We train with the AdamW~\cite{loshchilov2017adamw} optimizer for 90 epochs at a learning rate of $1\times 10^{-4}$ and a weight decay of $5\times 10^{-3}$.
The loss weights are set as $\lambda_{\mathrm{ray}} = \lambda_{\mathrm{color}} = \lambda_{\mathrm{event}} = \lambda_{\mathrm{o}} = 1.0$, $\lambda_{\mathrm{d}} = 0.1$, $\lambda_{\mathrm{SSIM}} = 0.2$, and $\lambda_{\mathrm{LPIPS}} = 0.1$.
The predicted color at query time $t$ is parameterized as a residual over the linear blend $(1-t)\,I_s + t\,I_e$ of the endpoint frames.
We scale the color loss by $50\times$ at pixels where an event is observed, to focus more on regions where the appearance actually changes.

\paragraph{Baselines.}
We compare against interpolation methods at both ends of the geometry pipeline, input-space and output-space.
As input-space interpolation baselines, which feed event-based RGB interpolation outputs into a frame-based geometric foundation model, we use TimeLens~\cite{tulyakov2021time}, a representative event-based continuous-time RGB interpolator, and VDM-EVFI~\cite{chen2025repurposing}, a more recent diffusion-based interpolator.
Their synthesized intermediate frames are passed to the frozen DA3 backbone for depth and camera pose prediction.
As the output-space interpolation baseline, we use Interp3R~\cite{guo2026interp3r}, which interpolates pointmaps from Align3R~\cite{lu2025align3r}, initializes its decoder from Align3R, and is trained only on PointOdyssey~\cite{zheng2023pointodyssey}.
Since Interp3R is not open-sourced, we reimplement it for a fair comparison.
Our reimplementation is built on DA3, takes the DA3 point cloud as input, initializes its decoder from DA3, and is trained on the same datasets and for the same number of epochs as our model.
We additionally include an \textit{Ours w/o events} variant that shares all components with our full model but removes the event-based encoder in the context feature extraction module in~\cref{subsec:network}, performing latent feature interpolation from RGB inputs alone.
All other training settings are kept identical to the full model.

\paragraph{Datasets.}
We evaluate on the real-world event dataset DSEC~\cite{gehrig2021dsec} and the synthetic dataset Sim4D~\cite{matsuki20254dtam}.
DSEC is an RGB-event driving dataset providing ground-truth depth, captured with a Prophesee Gen3.1 event camera synchronized with RGB cameras and a LiDAR.
For ground-truth camera poses, we follow ES-PTAM~\cite{ghosh2024ptam} and use LiDAR-inertial odometry as reference.
For evaluation, we take the first 200 frames of the \textit{zurich\_city\_04\_a-d} sequences.
Sim4D~\cite{matsuki20254dtam} is a Blender-based synthetic dataset of moving objects and cameras, providing accurate ground-truth depth and camera pose at a minimum of 120 FPS.
From its high-FPS frames, we synthesize events~\cite{gehrig2020video}, motion blur, and low-light images~\cite{bae20242,liu2021mba,bae2026roel} for the robustness analysis in \cref{subsec:robustness}.
We use the \textit{curtain}, \textit{modular\_vehicle}, \textit{rhino}, \textit{mercedes}, and \textit{wave\_toy} scenes.
For both DSEC and Sim4D, we sample frame pairs at query intervals of 2, 4, 8, and 16 frames, following previous works~\cite{tulyakov2021time,chen2025repurposing,guo2026interp3r}.
An interval of $T$ corresponds to $T-1$ intermediate frames between the endpoints $I_s$ and $I_e$.
For training, we synthesize pseudo GT supervision (\cref{subsec:training}) from two high-frame-rate RGB datasets, Adobe240~\cite{su2017deep} and SportsSloMo~\cite{chen2023sportsslomo}, both of which provide 240FPS real-world video.
We additionally synthesize motion blur from the high-frame-rate frames and use blurry inputs in 30\% of training data for all methods.
For Adobe240, we follow the standard training split of~\cite{xu2021temporal}.
For SportsSloMo, we use our own training/test split, and all qualitative results reported in the paper are drawn from this test set.
We additionally train on PointOdyssey~\cite{zheng2023pointodyssey}, the synthetic dataset used by Interp3R.
Since PointOdyssey does not provide high-frame-rate frames, its simulated events are markedly less accurate than other datasets.
In addition, a domain gap exists between this synthetic data and real-world scenes.
Our ablation study (\cref{subsec:ablation}) shows that adding PointOdyssey yields negligible improvement over our pseudo-GT supervision alone.

\paragraph{Metrics.}
For depth, we follow Interp3R~\cite{guo2026interp3r} and report the Absolute Relative Error (AbsRel) and the threshold accuracy $\delta < 1.25$; we additionally report the stricter thresholds $\delta < 1.05$ and $\delta < 1.10$ for a more informative comparison.
For pose, as in Interp3R, we report the Absolute Trajectory Error (ATE), Relative Translation Error (RTE), and Relative Rotation Error (RRE), computed with the \texttt{evo}~\cite{grupp2017evo} library.
Both depth and camera pose are evaluated over the ground-truth frames from $t = 0$ to $t = 1$.
For appearance interpolation, we report PSNR, SSIM, and VGG-based LPIPS~\cite{simonyan2014very, zhang2018unreasonable}.
We exclude the endpoint frames ($t = 0$ and $t = 1$) from the color metrics, since for sharp inputs their ground-truth colors are identical to the input frames $I_s$ and $I_e$ themselves.

\begin{table*}[t]
  \centering
  \small
  \setlength{\tabcolsep}{4pt}
  \resizebox{0.9\textwidth}{!}{%
  \begin{tabular}{c l c c c c c c c}
    \toprule
    \multirow{2}{*}{Interval} & \multirow{2}{*}{Method} & \multicolumn{4}{c}{Depth} & \multicolumn{3}{c}{Camera Pose} \\
    \cmidrule(lr){3-6} \cmidrule(lr){7-9}
    & & Abs Rel $\downarrow$ & $\delta < 1.05$ $\uparrow$ & $\delta < 1.10$ $\uparrow$ & $\delta < 1.25$ $\uparrow$ & ATE $\downarrow$ & RTE $\downarrow$ & RRE $\downarrow$ \\
    \midrule
    \multirow{5}{*}{2}
      & TimeLens      & 0.1462          & 0.3195          & 0.5381          & 0.8233          & \textbf{0.0471} & 1.1440           & \textbf{0.7962} \\
      & VDM-EVFI      & 0.1415          & 0.3179          & 0.5379          & 0.8261          & 0.2873          & 1.1709          & 0.9030           \\
      & Interp3R      & 0.1409          & 0.3254          & 0.5503          & 0.8393          & 0.2916          & 1.2992          & 1.9502          \\
      & w/o events    & 0.1441          & 0.3370           & 0.5486          & 0.8329          & 0.1503          & 1.1490           & 1.0245          \\
      & Ours & \textbf{0.1355} & \textbf{0.3428} & \textbf{0.5603} & \textbf{0.8457} & 0.0700            & \textbf{1.1184} & 0.9067          \\
    \midrule
    \multirow{5}{*}{4}
      & TimeLens      & 0.1546          & 0.3114          & 0.5266          & 0.8110          & 0.1732 & 1.1491          & \textbf{0.7845} \\
      & VDM-EVFI      & 0.1549          & 0.3020           & 0.5158          & 0.7980           & 0.5011          & 1.2320          & 0.8526          \\
      & Interp3R      & 0.1537          & 0.2960           & 0.5207          & 0.8199          & 0.5839          & 1.3165          & 1.6952          \\
      & w/o events    & 0.1635          & 0.3038          & 0.5118          & 0.8013          & 0.2698          & 1.1673          & 0.9678          \\
      & Ours & \textbf{0.1428} & \textbf{0.3337} & \textbf{0.5502} & \textbf{0.8344} & \textbf{0.1399}          & \textbf{1.1413} & 0.9116          \\
    \midrule
    \multirow{5}{*}{8}
      & TimeLens      & 0.1945          & 0.2610           & 0.4638          & 0.7478          & 0.5342          & 1.2162          & 0.9908          \\
      & VDM-EVFI      & 0.1783          & 0.2628          & 0.4635          & 0.7615          & 0.4729          & 1.2053          & \textbf{0.8357} \\
      & Interp3R      & 0.1763          & 0.2441          & 0.4534          & 0.7781          & 1.1725          & 1.4478          & 1.4771          \\
      & w/o events    & 0.1959          & 0.2602          & 0.4486          & 0.7384          & 0.4264 & \textbf{1.1631} & 0.8372          \\
      & Ours & \textbf{0.1628} & \textbf{0.3056} & \textbf{0.5070}  & \textbf{0.7917} & \textbf{0.2819}          & 1.1650           & 0.8806          \\
    \midrule
    \multirow{5}{*}{16}
      & TimeLens      & 0.3010           & 0.1661          & 0.3149          & 0.5907          & 1.7852          & 1.5321          & 6.9506          \\
      & VDM-EVFI      & 0.2278          & 0.1969          & 0.3645          & 0.6611          & 0.5508          & 1.2257          & \textbf{0.8593} \\
      & Interp3R      & 0.2071          & 0.1959          & 0.3718          & 0.7140           & 2.1770           & 1.6286          & 1.2675          \\
      & w/o events    & 0.2302          & 0.2065          & 0.3761          & 0.6779          & 0.7423          & 1.2475          & 0.9940           \\
      & Ours & \textbf{0.1965} & \textbf{0.2507} & \textbf{0.4406} & \textbf{0.7308} & \textbf{0.4050}  & \textbf{1.2096} & 0.9757          \\
    \midrule
    \multirow{5}{*}{Avg}
      & TimeLens      & 0.1991          & 0.2645          & 0.4609          & 0.7432          & 0.6349          & 1.2604          & 2.3805          \\
      & VDM-EVFI      & 0.1756          & 0.2699          & 0.4704          & 0.7617          & 0.4530           & 1.2085          & \textbf{0.8627} \\
      & Interp3R      & 0.1695          & 0.2653          & 0.4741          & 0.7878          & 1.0563          & 1.4230           & 1.5975          \\
      & w/o events    & 0.1834          & 0.2769          & 0.4713          & 0.7626          & 0.3972          & 1.1817          & 0.9559          \\
      & Ours & \textbf{0.1594} & \textbf{0.3082} & \textbf{0.5145} & \textbf{0.8007} & \textbf{0.2242} & \textbf{1.1586} & 0.9187          \\
    \bottomrule
  \end{tabular}
  }
  \caption{Quantitative comparison on DSEC.
    We compare against input-space interpolation baselines (TimeLens, VDM-EVFI) and an output-space interpolation baseline (Interp3R) across query intervals of 2, 4, 8, and 16 frames, and report the average over all intervals.
    The best result in each column is marked in bold.}
  \vspace{-1.0em}
  \label{tab:dsec}
\end{table*}

\begin{figure*}[t]
  \centering
  \setlength{\tabcolsep}{0.5pt}
  \renewcommand{\arraystretch}{0.5}
  \small
  \begin{tabular}{ccccc}
    GT RGB & TimeLens & Interp3R & Ours & GT Depth \\
    \includegraphics[width=0.198\linewidth]{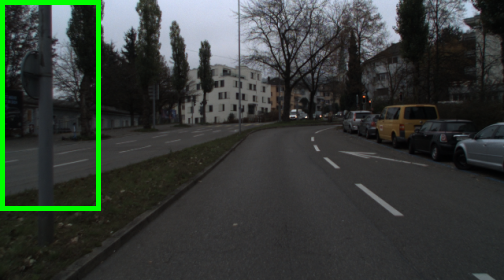} &
    \includegraphics[width=0.198\linewidth]{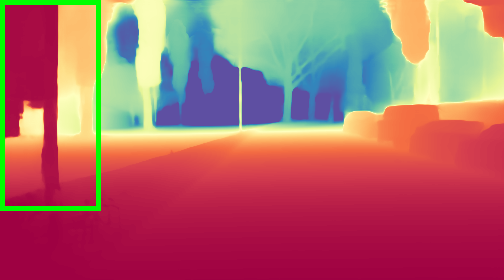} &
    \includegraphics[width=0.198\linewidth]{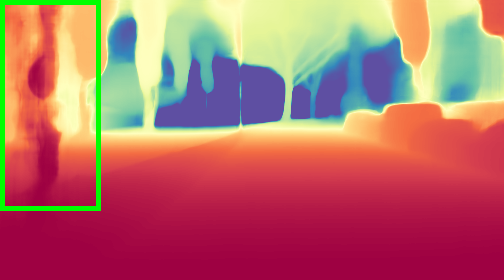} &
    \includegraphics[width=0.198\linewidth]{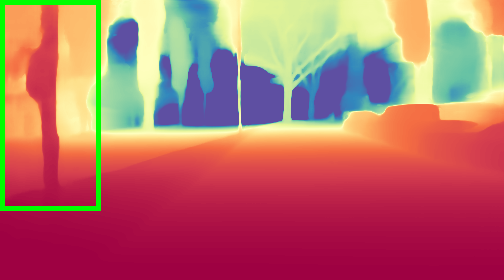} &
    \includegraphics[width=0.198\linewidth]{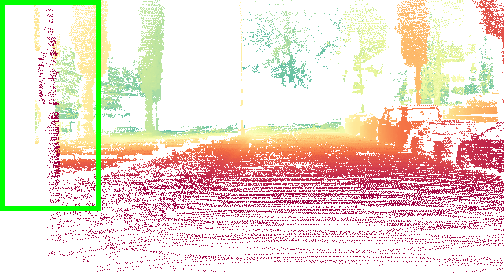} \\
    \includegraphics[width=0.198\linewidth]{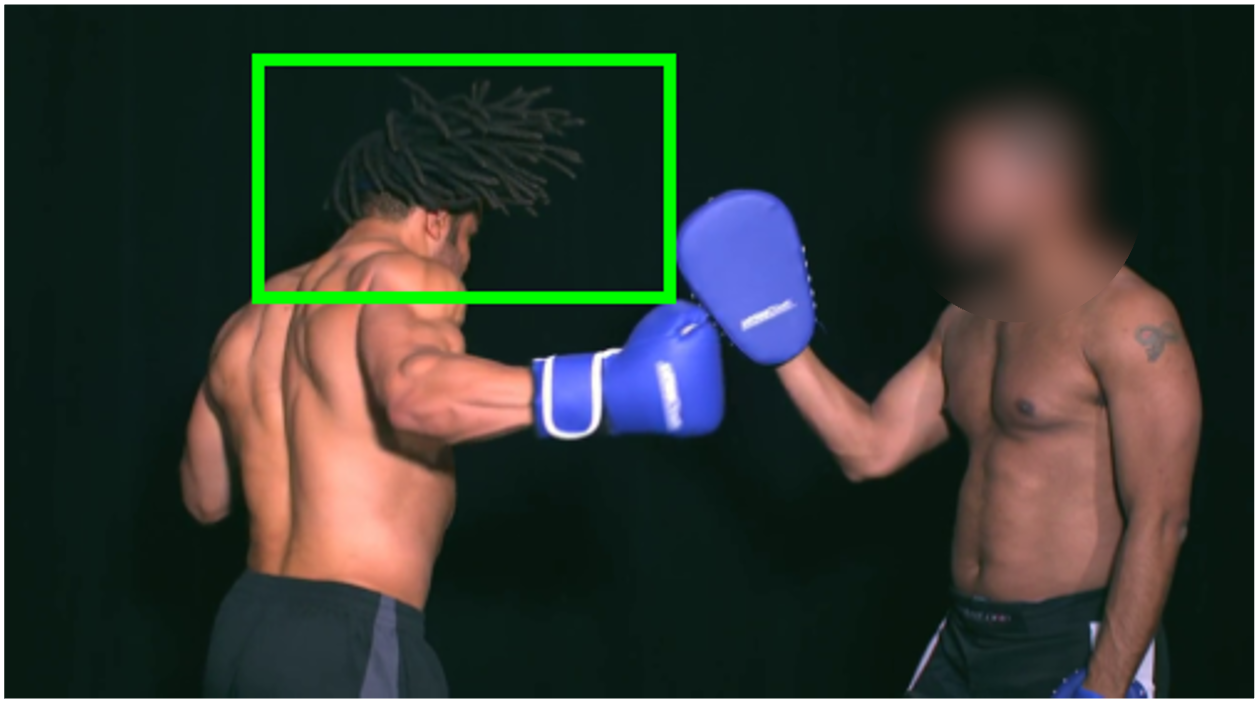} &
    \includegraphics[width=0.198\linewidth]{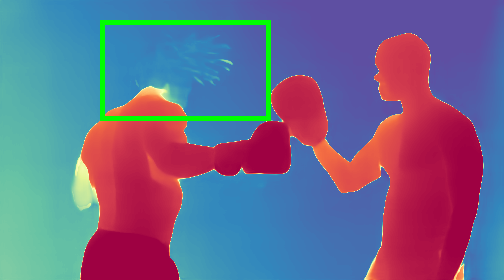} &
    \includegraphics[width=0.198\linewidth]{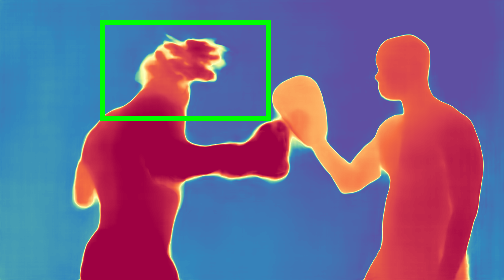} &
    \includegraphics[width=0.198\linewidth]{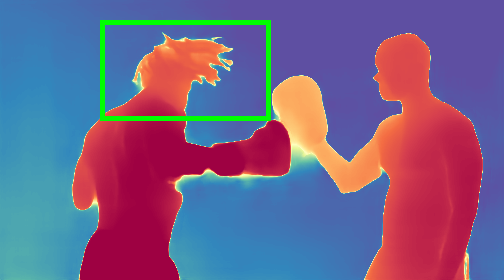} &
    \includegraphics[width=0.198\linewidth]{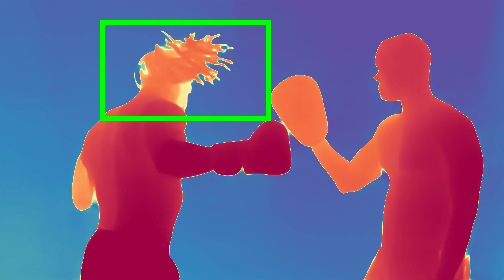} \\
    \includegraphics[width=0.198\linewidth]{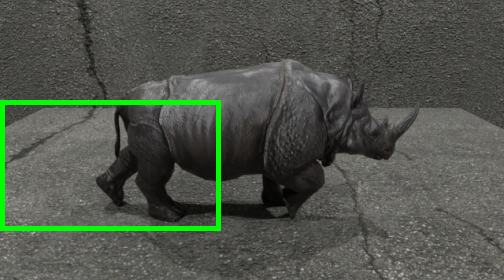} &
    \includegraphics[width=0.198\linewidth]{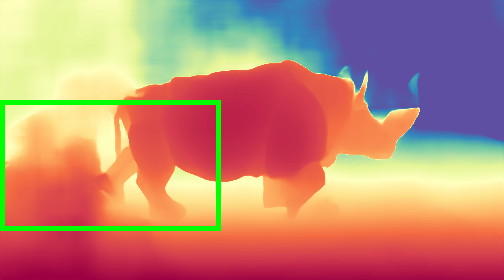} &
    \includegraphics[width=0.198\linewidth]{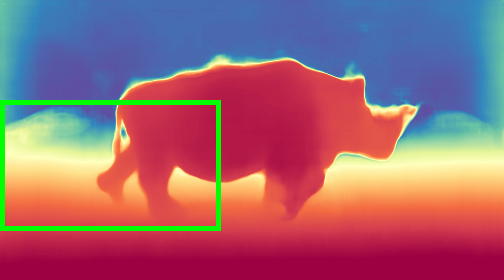} &
    \includegraphics[width=0.198\linewidth]{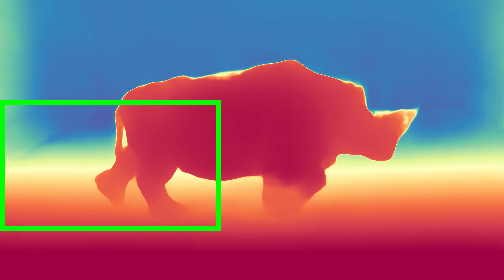} &
    \includegraphics[width=0.198\linewidth]{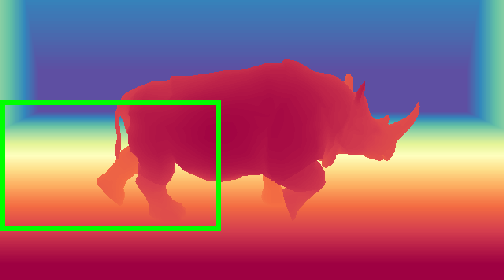} \\
  \end{tabular}
  \caption{Qualitative depth-interpolation comparison at $t = 0.5$ on DSEC
    (top), SportsSloMo (middle), and Sim4D (bottom). Green boxes mark the
    region of interest in each row.}
  \vspace{-1.3em}
  \label{fig:depth_comparison}
\end{figure*}

\subsection{Comparison with Baselines}
\cref{tab:dsec} reports the quantitative results on the real-world DSEC dataset.
Our method shows superior depth interpolation at every interval, outperforming all baselines.
This shows that the model trained with our self-supervised framework transfers effectively to real-world events.
The gap between our full model and the \textit{Ours w/o events} variant further indicates that the event stream is being effectively used as a temporal cue.
For pose interpolation, TimeLens attains a lower ATE at interval 2, but our method overtakes it at the more challenging longer intervals and yields the best average ATE overall.

\Cref{fig:depth_comparison} presents the qualitative comparison.
TimeLens accumulates errors through its two-stage pipeline of RGB interpolation followed by geometry estimation.
On DSEC, regions around poles are reconstructed as sharp yet inaccurate, while on SportsSloMo and Sim4D, geometry errors arise where the foreground have similar colors to the background.
These failures indicate that depth interpolation errors grow where RGB interpolation itself is most difficult.
Interp3R, an output-space pointmap interpolator, recovers more accurate depth than TimeLens but produces blurry boundaries.
In contrast, our latent-space interpolation yields both sharp and accurate boundaries.
Qualitative results of VDM-EVFI and \textit{Ours w/o events} can be found in the supplementary material.

\subsection{Robustness to Corruptions}
\label{subsec:robustness}
\begin{table*}[t]
  \centering
  \small
  \setlength{\tabcolsep}{4pt}
  \resizebox{0.85\textwidth}{!}{
  \begin{tabular}{c l c c c c c c c}
    \toprule
    \multirow{2}{*}{Data} & \multirow{2}{*}{Method} & \multicolumn{4}{c}{Depth} & \multicolumn{3}{c}{Camera Pose} \\
    \cmidrule(lr){3-6} \cmidrule(lr){7-9}
    & & Abs Rel $\downarrow$ & $\delta < 1.05$ $\uparrow$ & $\delta < 1.10$ $\uparrow$ & $\delta < 1.25$ $\uparrow$ & ATE $\downarrow$ & RTE $\downarrow$ & RRE $\downarrow$ \\
    \midrule
    \multirow{5}{*}{Sharp}
      & TimeLens      & 0.0216         & 0.9282          & 0.9664          & 0.9857          & \textbf{0.0097} & 0.0594          & 1.8919          \\
      & VDM-EVFI      & 0.0333         & 0.8332          & 0.9333          & 0.9826          & 0.0205          & 0.0570          & 1.7865          \\
      & Interp3R      & 0.0215         & 0.9323          & 0.9716          & 0.9868          & 0.0312          & \textbf{0.0358} & \textbf{1.4278} \\
      & w/o events    & 0.0243         & 0.9182          & 0.9683          & 0.9855          & 0.0209          & 0.0521          & 1.7744          \\
      & Ours & \textbf{0.0160} & \textbf{0.9543} & \textbf{0.9826} & \textbf{0.9934} & 0.0208          & 0.0529          & 1.7343          \\
    \midrule
    \multirow{5}{*}{Blur}
      & TimeLens      & 0.0579          & 0.6981          & 0.8428          & 0.9408          & \textbf{0.0177} & 0.0560          & 1.3889          \\
      & VDM-EVFI      & 0.0585          & 0.6729          & 0.8339          & 0.9513          & 0.0190           & 0.0562          & \textbf{1.3170}  \\
      & Interp3R      & 0.0492          & 0.7379          & 0.8760          & 0.9557          & 0.0330          & \textbf{0.0365} & 1.3672          \\
      & w/o events    & 0.0549          & 0.6980           & 0.8715          & 0.9554          & 0.0240          & 0.0501          & 1.5527          \\
      & Ours & \textbf{0.0408} & \textbf{0.7747} & \textbf{0.9064} & \textbf{0.9736} & 0.0231          & 0.0512          & 1.5160          \\
    \midrule
    \multirow{5}{*}{Dark}
      & TimeLens      & 0.0901          & 0.5390          & 0.7286          & 0.8824          & 0.0255          & 0.0567          & 1.5778          \\
      & VDM-EVFI      & 0.1582          & 0.3665          & 0.5265          & 0.7403          & 0.0343          & 0.0555          & 1.6237          \\
      & Interp3R      & 0.0664          & 0.6040          & 0.7986          & 0.9494          & 0.0323          & \textbf{0.0356} & \textbf{1.4747} \\
      & w/o events    & 0.0558          & 0.6727          & 0.8565          & 0.9618          & 0.0232          & 0.0515          & 1.6362          \\
      & Ours & \textbf{0.0492} & \textbf{0.7189} & \textbf{0.8781} & \textbf{0.9701} & \textbf{0.0226} & 0.0520           & 1.6093          \\
    \bottomrule
  \end{tabular}
  }
  \caption{Robustness analysis on Sim4D.
    Depth and pose interpolation performance under sharp, motion-blur, and low-light (dark) conditions.
    The best results are marked in bold.}
  \vspace{-0.5em}
  \label{tab:sim4d}
\end{table*}

\begin{figure*}[t]
  \centering
  \setlength{\tabcolsep}{0.5pt}
  \renewcommand{\arraystretch}{0.5}
  \small
  \newcommand{\rowlabel}[1]{%
    \adjustbox{valign=c}{\rotatebox{90}{#1}}}
  \newcommand{\fig}[1]{%
    \includegraphics[width=0.198\linewidth,valign=c]{#1}}
\resizebox{0.85\textwidth}{!}{%
  \begin{tabular}{cccccc}
    & Input Keyframe & TimeLens & Interp3R & Ours & GT \\
    \rowlabel{Sharp} &
    \fig{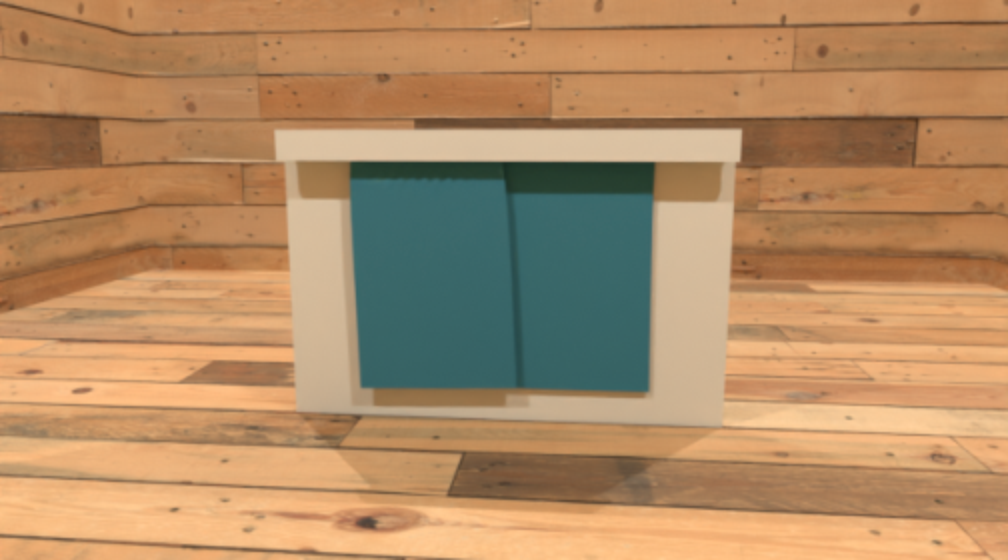} &
    \fig{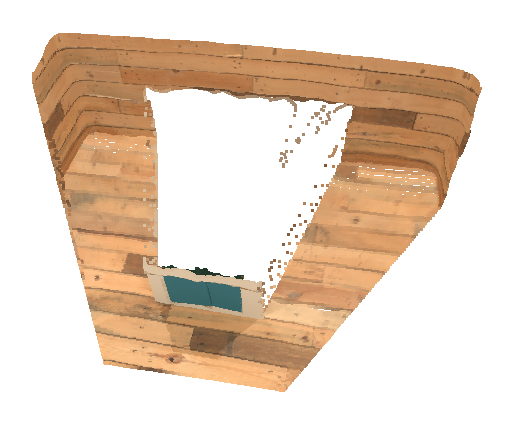} &
    \fig{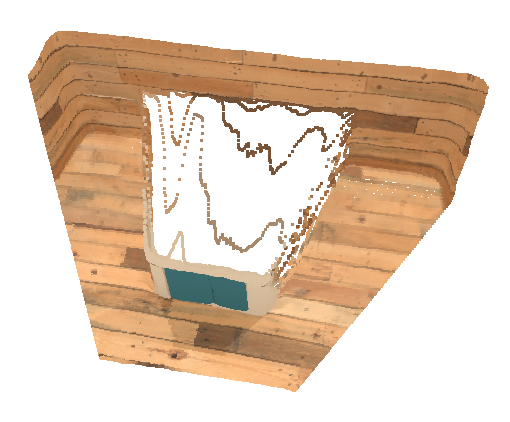} &
    \fig{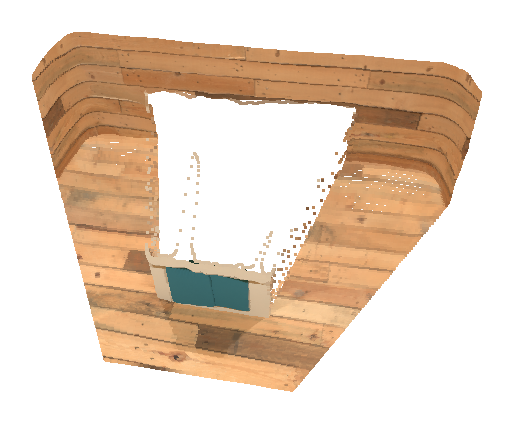} &
    \fig{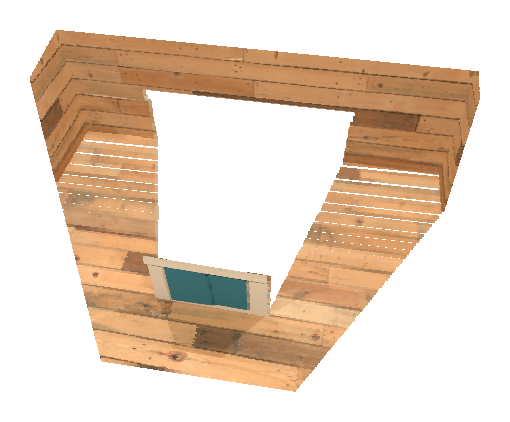} \\
    \rowlabel{Blur} &
    \fig{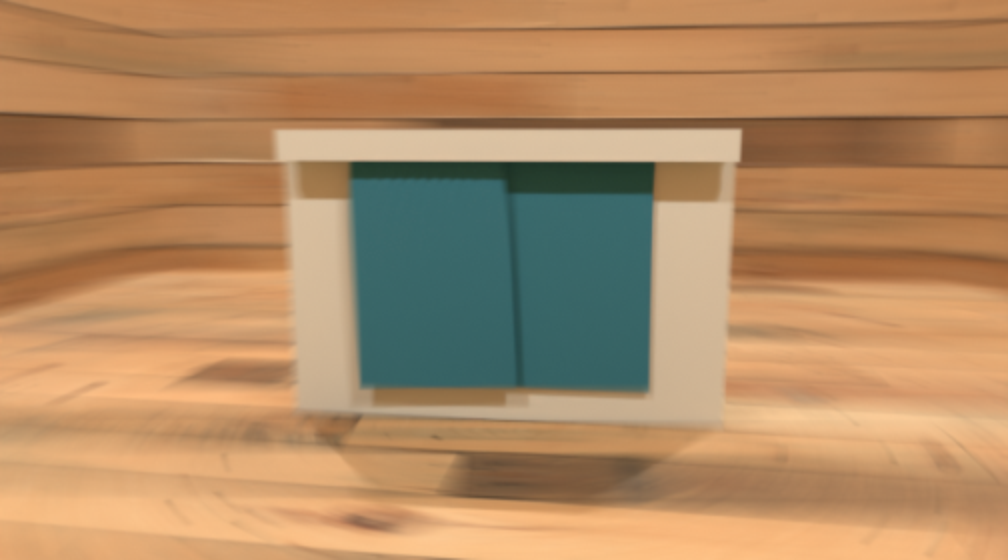} &
    \fig{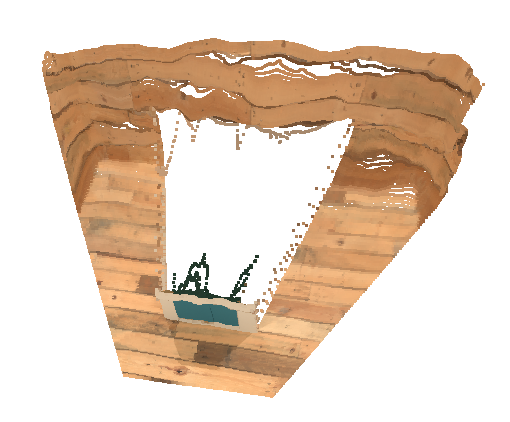} &
    \fig{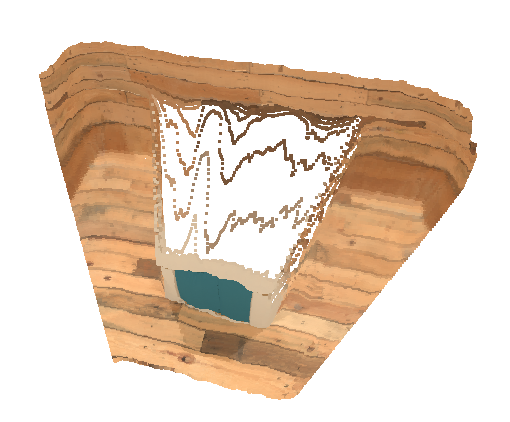} &
    \fig{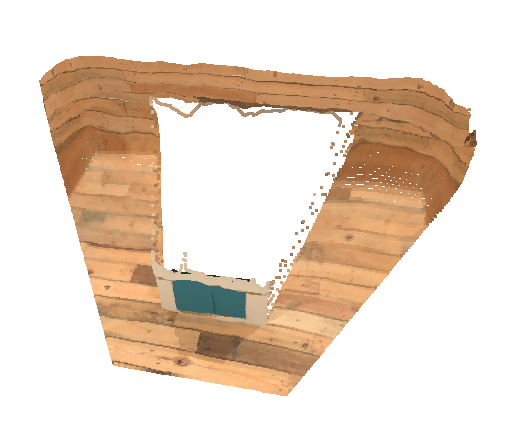} &
    \fig{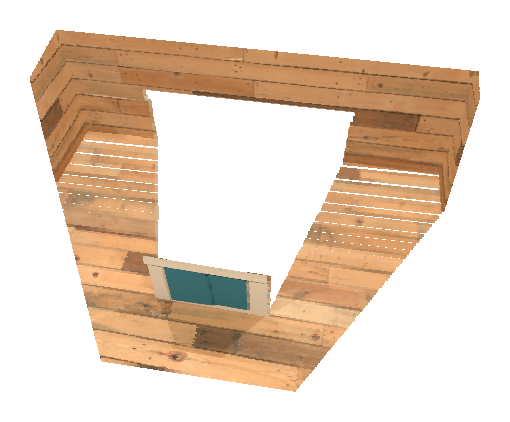} \\
    \rowlabel{Dark} &
    \fig{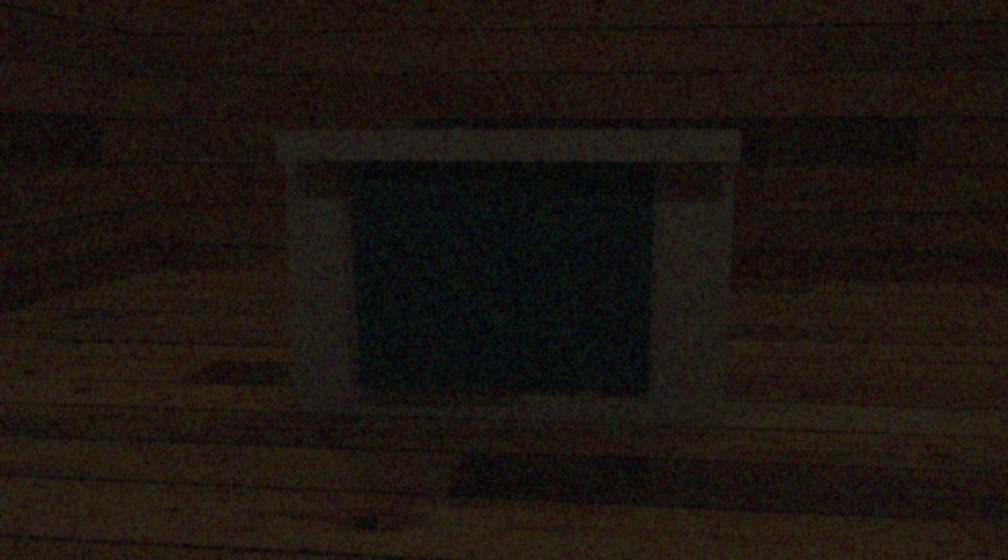} &
    \fig{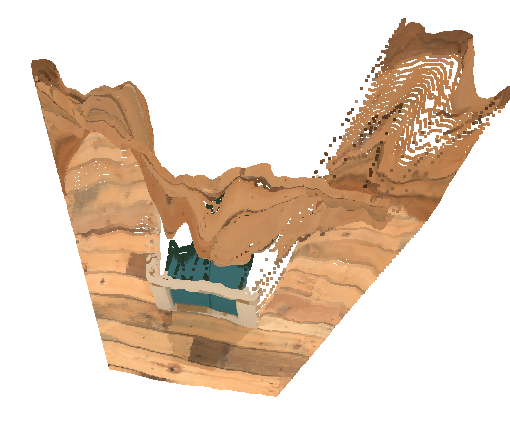} &
    \fig{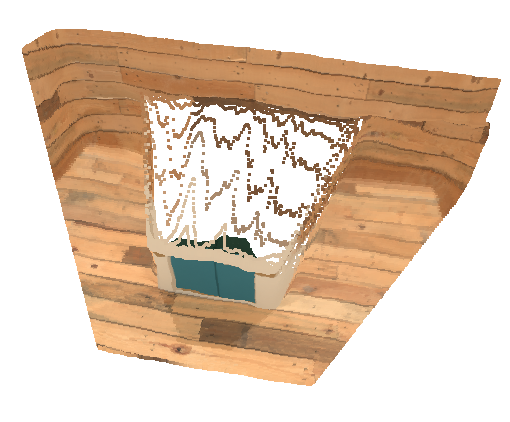} &
    \fig{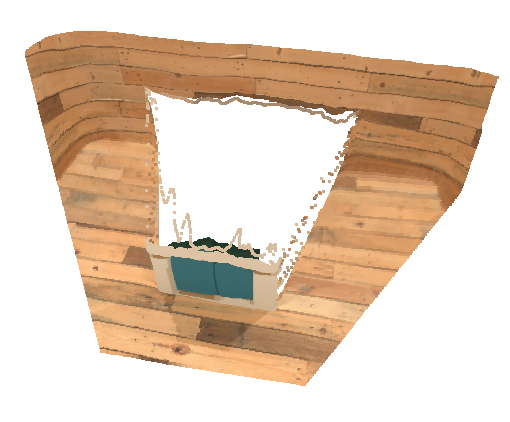} &
    \fig{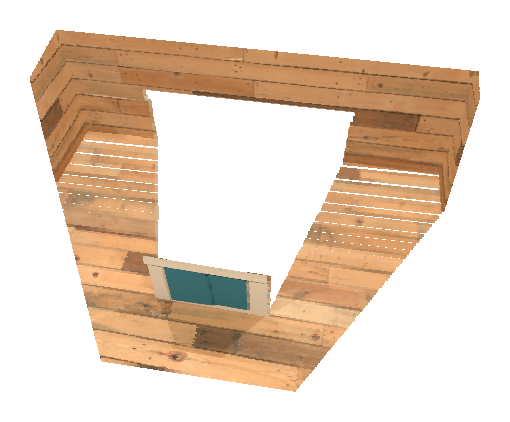} \\
  \end{tabular}
  }
  \vspace{0.0em}
  \caption{Robustness analysis under synthetic challenging imaging conditions
    on the Sim4D dataset. From top to bottom: sharp, blur, and low-light inputs. The Input Keyframe column shows the input frame at $t = 0$.
    All point cloud results are colored with the GT RGB image.}
  \vspace{-1.2em}
  \label{fig:robustness}
\end{figure*}

We next analyze each method under sharp, motion-blur, and low-light imaging conditions, both quantitatively and qualitatively, using Sim4D~\cite{matsuki20254dtam}.
\Cref{tab:sim4d} shows the depth and pose interpolation performance under sharp, motion-blur, and low-light imaging conditions.
This table reports the results averaged over all intervals. Per-interval results are provided in the supplementary material.
Our method achieves the strongest depth interpolation in every condition.
The degradation is particularly severe for the input-space RGB interpolators (TimeLens and VDM-EVFI) under low light, where the corrupted input directly causes errors in the predicted geometry.
On Absolute Trajectory Error, our approach consistently surpasses Interp3R across all conditions, although it is worse than the image-space methods on sharp and blurry inputs.
It is likely because the simple scene structure of Sim4D favors image-space pipelines when the input is clean.
Under low-light condition, however, the image-space baselines break down while our model continues to deliver accurate pose estimates.
Interp3R attains the strongest relative translation accuracy (RTE) but performs poorly on the ATE, which captures global trajectory accuracy.

The qualitative comparison in \cref{fig:robustness} supports these trends.
TimeLens produces large geometry errors under motion-blur, and entirely fails to separate the foreground object from the background in low-light scenes.
Interp3R is more robust than TimeLens overall but introduces visible boundary artifacts.
In contrast, our method preserves the foreground-background separation even under motion-blur and low-light.

\subsection{Appearance Interpolation}
\begin{table*}[t]
  \centering
  \small
  \setlength{\tabcolsep}{4pt}
  \resizebox{0.85\textwidth}{!}{
  \begin{tabular}{c l c c c c c c c}
    \toprule
    \multirow{2}{*}{Data} & \multirow{2}{*}{Method} & \multicolumn{3}{c}{Color} & \multicolumn{4}{c}{Depth} \\
    \cmidrule(lr){3-5} \cmidrule(lr){6-9}
    & & PSNR $\uparrow$ & SSIM $\uparrow$ & LPIPS $\downarrow$ & Abs Rel & $\delta < 1.05$ & $\delta < 1.10$ & $\delta < 1.25$ \\
    \midrule
    \multirow{3}{*}{DSEC}
      & TimeLens      & 23.97          & \textbf{0.7175} & 0.3573          & 0.1991          & 0.2645          & 0.4609          & 0.7432          \\
      & VDM-EVFI      & 21.85          & 0.6724          & 0.4439          & 0.1756          & 0.2699          & 0.4704          & 0.7617          \\
      & Ours & \textbf{24.18} & 0.7168          & \textbf{0.3478} & \textbf{0.1594} & \textbf{0.3082} & \textbf{0.5145} & \textbf{0.8007} \\
    \midrule
    \multirow{3}{*}{\shortstack{Sim4D \\ Sharp}}
      & TimeLens      & \textbf{32.44} & \textbf{0.8412} & \textbf{0.1678} & 0.0216         & 0.9282          & 0.9664          & 0.9857          \\
      & VDM-EVFI      & 27.67          & 0.7371          & 0.2883          & 0.0333         & 0.8332          & 0.9333          & 0.9826          \\
      & Ours          & 29.97          & 0.7484          & 0.2158          & \textbf{0.0160} & \textbf{0.9543} & \textbf{0.9826} & \textbf{0.9934} \\
    \midrule
    \multirow{3}{*}{\shortstack{Sim4D \\ Blur}}
      & TimeLens      & 28.64         & 0.7488          & 0.3700          & 0.0579          & 0.6981          & 0.8428          & 0.9408          \\
      & VDM-EVFI      & 26.80         & 0.7215          & 0.4258          & 0.0585          & 0.6729          & 0.8339          & 0.9513          \\
      & Ours & \textbf{29.16} & \textbf{0.7772} & \textbf{0.3436} & \textbf{0.0408} & \textbf{0.7747} & \textbf{0.9064} & \textbf{0.9736} \\
    \bottomrule
  \end{tabular}
  }
  \caption{Appearance interpolation analysis.
    Color and depth interpolation performance of our method compared with event-based color interpolation baselines, TimeLens and VDM-EVFI, on DSEC, Sim4D-Sharp, and Sim4D-Blur.
    The best result in each column is in bold.}
  \vspace{-0.7em}
  \label{tab:proxy}
\end{table*}

We next examine appearance interpolation quality and its relationship with geometry accuracy.
\cref{tab:proxy} reports color and depth errors on DSEC, Sim4D-Sharp, and Sim4D-Blur.
Interp3R is omitted as it does not produce color interpolations.
Although our model is not specialized for color synthesis, it still shows the best PSNR on the real-world dataset.
On the clean synthetic data, Sim4D-Sharp, TimeLens substantially outperforms our method on color metrics.
Even in this regime, however, our depth interpolation remains the most accurate, indicating that gains in color quality do not directly translate into better depth.
On the blurry synthetic data, Sim4D-Blur, our approach surpasses TimeLens on color interpolation as well, showing robustness of our method.

\subsection{Continuous-time Interpolation}
\begin{figure*}[t]
  \centering
  \setlength{\tabcolsep}{1pt}
  \renewcommand{\arraystretch}{0.5}
  \small
  \resizebox{0.8\textwidth}{!}{
  \begin{tabular}{ccccc}
    $t = 0.0$ & $t = 0.25$ & $t = 0.5$ & $t = 0.75$ & $t = 1.0$ \\
    \includegraphics[width=0.198\linewidth]{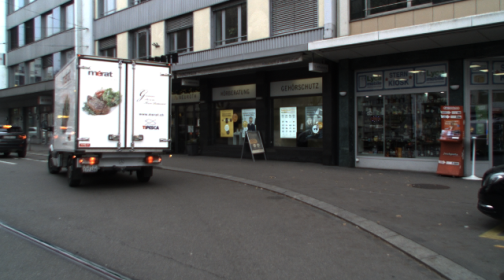} &
    \includegraphics[width=0.198\linewidth]{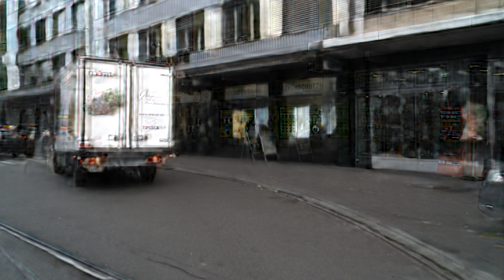} &
    \includegraphics[width=0.198\linewidth]{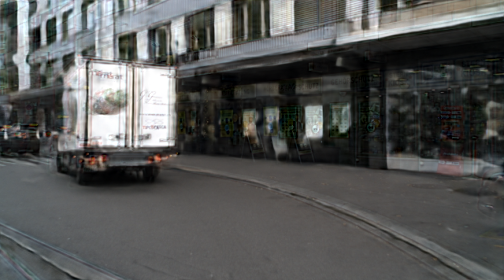} &
    \includegraphics[width=0.198\linewidth]{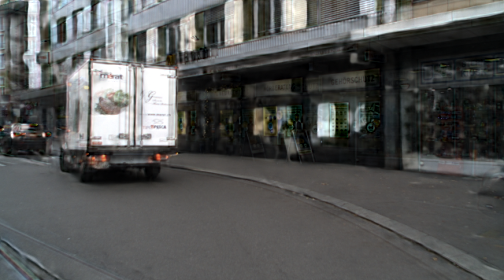} &
    \includegraphics[width=0.198\linewidth]{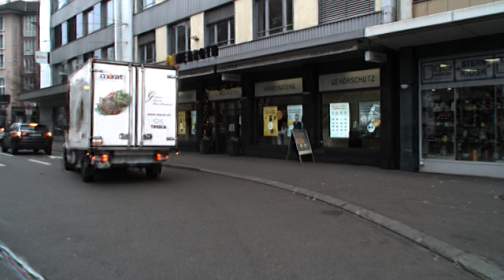} \\
    \includegraphics[width=0.198\linewidth]{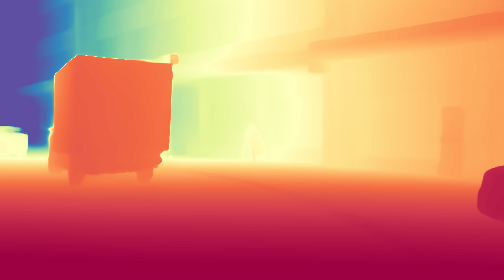} &
    \includegraphics[width=0.198\linewidth]{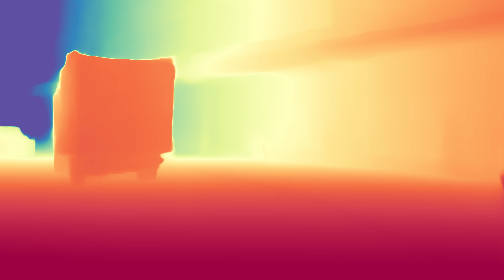} &
    \includegraphics[width=0.198\linewidth]{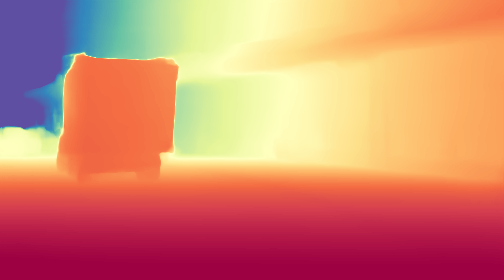} &
    \includegraphics[width=0.198\linewidth]{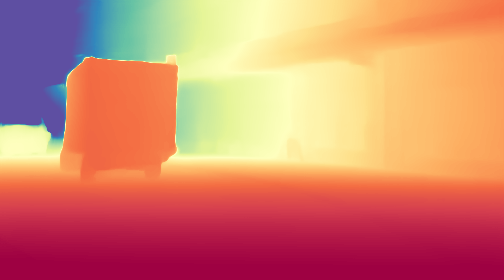} &
    \includegraphics[width=0.198\linewidth]{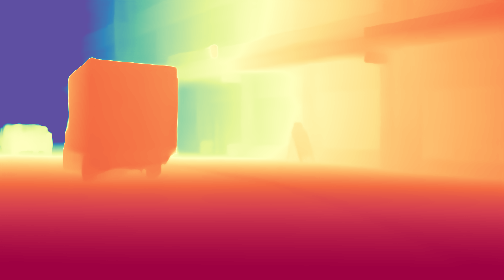} \\
    \includegraphics[width=0.198\linewidth]{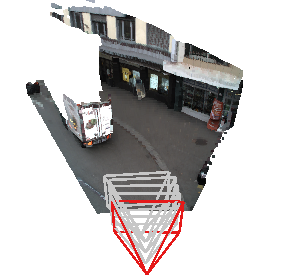} &
    \includegraphics[width=0.198\linewidth]{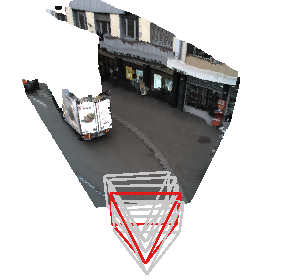} &
    \includegraphics[width=0.198\linewidth]{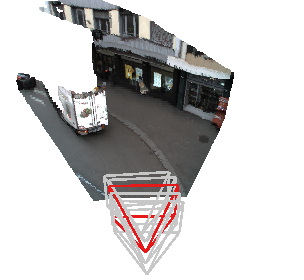} &
    \includegraphics[width=0.198\linewidth]{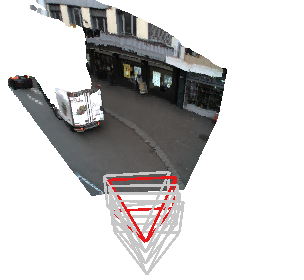} &
    \includegraphics[width=0.198\linewidth]{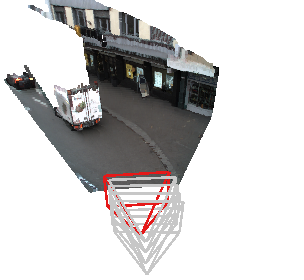} \\
  \end{tabular}
  }
  \vspace{0em}
  \caption{Continuous prediction of our method across $t \in [0.0, 1.0]$ on a DSEC sequence. From top to bottom, the rows show the predicted RGB, depth, and point cloud. The red frustum marks the predicted camera pose at the current timestep.}
  \vspace{-1.5em}
  \label{fig:continuous}
\end{figure*}

We finally illustrate the continuous-time interpolation capability of our framework.
Because the appearance interpolation proxy is jointly trained alongside the geometric modules, our model simultaneously produces depth, pose, and color predictions at any query timestamp.
\cref{fig:continuous} visualizes color, depth, and pose predictions on a DSEC sequence.
Although this scene is challenging because the scene follows a moving truck, the pose interpolation still succeeds and the outputs stay temporally consistent.
For clarity, the figure displays a discrete set of query times, but our method can be evaluated at any timestamp within the interval.

\subsection{Ablation study}
\label{subsec:ablation}

\begin{table*}[t]
  \centering
  \small
  \setlength{\tabcolsep}{4pt}
  \resizebox{0.85\textwidth}{!}{
  \begin{tabular}{l c c c c c c c}
    \toprule
    & \multicolumn{4}{c}{Depth} & \multicolumn{3}{c}{Camera Pose} \\
    \cmidrule(lr){2-5} \cmidrule(lr){6-8}
    & Abs Rel $\downarrow$ & $\delta < 1.05$ $\uparrow$ & $\delta < 1.10$ $\uparrow$ & $\delta < 1.25$ $\uparrow$ & ATE $\downarrow$ & RTE $\downarrow$ & RRE $\downarrow$ \\
    \midrule
    w/o event input & 0.0247 & 0.9170 & 0.9696 & 0.9861 & 0.0212 & 0.0533 & 1.7655 \\
    w/o proxy loss  & 0.0206 & 0.9359 & 0.9762 & 0.9902 & 0.0210 & 0.0513 & 1.7345 \\
    w/o inr         & 0.0197 & 0.9387 & 0.9781 & 0.9911 & 0.0213 & 0.0521 & 1.7208 \\
    synthetic GT    & 0.0206 & 0.9394 & 0.9779 & 0.9911 & 0.0267 & 0.0410 & 1.7692 \\
    pseudo GT       & 0.0181 & 0.9446 & 0.9790 & 0.9917 & 0.0211 & 0.0532 & 1.7464 \\
    Full            & 0.0186 & 0.9463 & 0.9800 & 0.9918 & 0.0208 & 0.0532 & 1.7306 \\
    \bottomrule
  \end{tabular}
  }
  \vspace{0.3em}
  \caption{Ablation study.
    Effect of each component on depth and pose interpolation accuracy.
    The first three rows ablate the event input, the appearance-based proxy loss, and the INR-based refinement.
    The last two rows compare synthetic and pseudo ground-truth supervision, isolating the effect of our self-supervised training framework.}
  \vspace{-1.0em}
  \label{tab:ablation}
\end{table*}

\cref{tab:ablation} reports the ablation study on depth and pose interpolation.
For efficiency, each ablated model is trained for 60 epochs on 50\% of the training data and evaluated on Sim4D-Sharp.
Removing the event input causes the largest drop in depth accuracy, showing that the event stream plays a crucial role in temporal interpolation.
The appearance-based proxy loss and the INR-based refinement also each lead to clear reductions in depth error.
We further analyze the effect of our self-supervised training framework by varying the training data.
Here, \textit{synthetic GT} denotes training only on PointOdyssey, while \textit{pseudo GT} denotes training only on the data generated by our self-supervised framework.
When the model is trained only on synthetic data, both depth and pose accuracy drop substantially.
This is likely due to the inaccurate event simulation from low-FPS RGB images and the domain gap of synthetic data.
In contrast, training only on our pseudo-GT data from high-FPS real-world RGB videos performs almost as well as using both sources together.
This shows that our self-supervised training framework plays a key role in producing accurate interpolation.

\section{Conclusion}

We present an event-guided latent feature interpolation framework for continuous-time 4D reconstruction that predicts depth and camera pose at arbitrary timestamps between RGB frames.
Unlike prior input-space and output-space approaches that interpolate either RGB frames or pointmaps, our method builds a continuous latent feature field directly within the latent space of a pretrained geometric foundation model.
This design fully leverages its learned geometric prior while letting events guide fine-grained temporal changes.
To regularize the feature interpolation, we introduce an appearance-based proxy task that jointly predicts continuous color and matches simulated events from it against the observed event stream.
We further propose a self-supervised training framework to overcome the scarcity of event datasets with ground-truth geometry.
Built on high-frame-rate RGB videos, this framework simulates realistic events and distills pseudo-ground-truth depth and camera pose from a foundation model teacher.
From extensive evaluation and analysis on real-world and synthetic corruptions, we demonstrate that our method outperforms RGB-interpolation and pointmap-interpolation baselines and remains robust where these baselines break down.
Our ablation study further highlights the effectiveness of the appearance-based proxy task and the self-supervised training framework.

\paragraph{Acknowledgement}
This work was partly supported by Institute of Information \& communications Technology Planning \& Evaluation (IITP) grant funded by the Korea government(MSIT) [RS-2021-II211343, Artificial Intelligence Graduate School Program and RS-2025-25442338, AI star Fellowship Support Program (Seoul National Univ.)], and Creative-Pioneering Researchers Program through Seoul National University.

\bibliography{egbib}

\clearpage
\appendix

\makeatletter
\begin{raggedright}
  {\bmv@titlesize\bfseries\sffamily\color{bmv@sectioncolor}
  Supplementary Material for\\
  Continuous-time 4D Reconstruction via Event-guided Latent Feature Interpolation\par}
\end{raggedright}
\vskip\baselineskip
\hrule
\vskip\baselineskip
\makeatother

\section{Additional Qualitative Comparison}
\label{sec:supp_qualitative}

We additionally conduct qualitative comparison against two baselines that are not shown in the main paper: \textit{VDM-EVFI}~\cite{chen2025repurposing} and \textit{Ours w/o events}.
VDM-EVFI~\cite{chen2025repurposing} is an event-based RGB interpolation model built on top of a diffusion backbone.
\textit{Ours w/o events} is an ablated variant of our model that does not use the event stream as input.
It learns latent feature interpolation solely from the two endpoint RGB frames.
For a direct comparison with other baselines, TimeLens~\cite{tulyakov2021time} and Interp3R~\cite{guo2026interp3r}, we report qualitative results on the same scenes as \cref{fig:depth_comparison}.

\begin{figure*}[t]
  \centering
  \setlength{\tabcolsep}{1pt}
  \renewcommand{\arraystretch}{0.5}
  \small
  \begin{tabular}{ccccc}
    GT RGB & VDM-EVFI & Ours w/o events & Ours & GT Depth \\
    \includegraphics[width=0.198\linewidth]{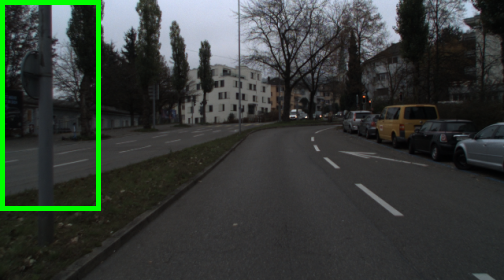} &
    \includegraphics[width=0.198\linewidth]{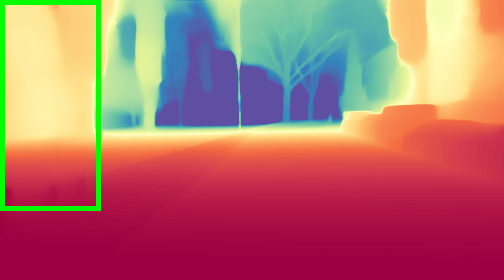} &
    \includegraphics[width=0.198\linewidth]{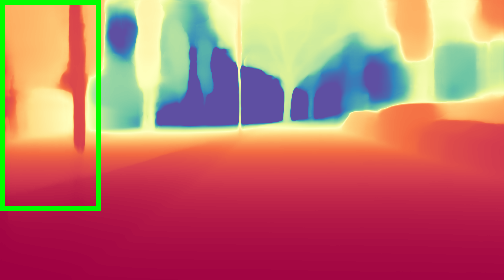} &
    \includegraphics[width=0.198\linewidth]{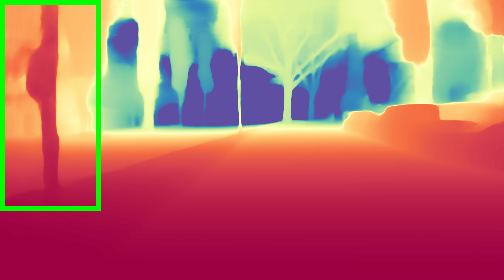} &
    \includegraphics[width=0.198\linewidth]{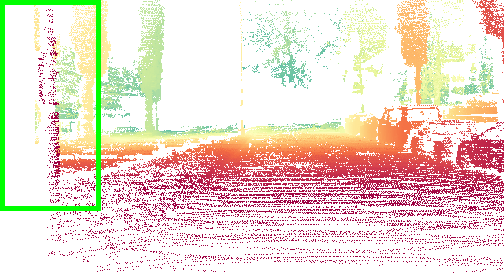} \\
    \includegraphics[width=0.198\linewidth]{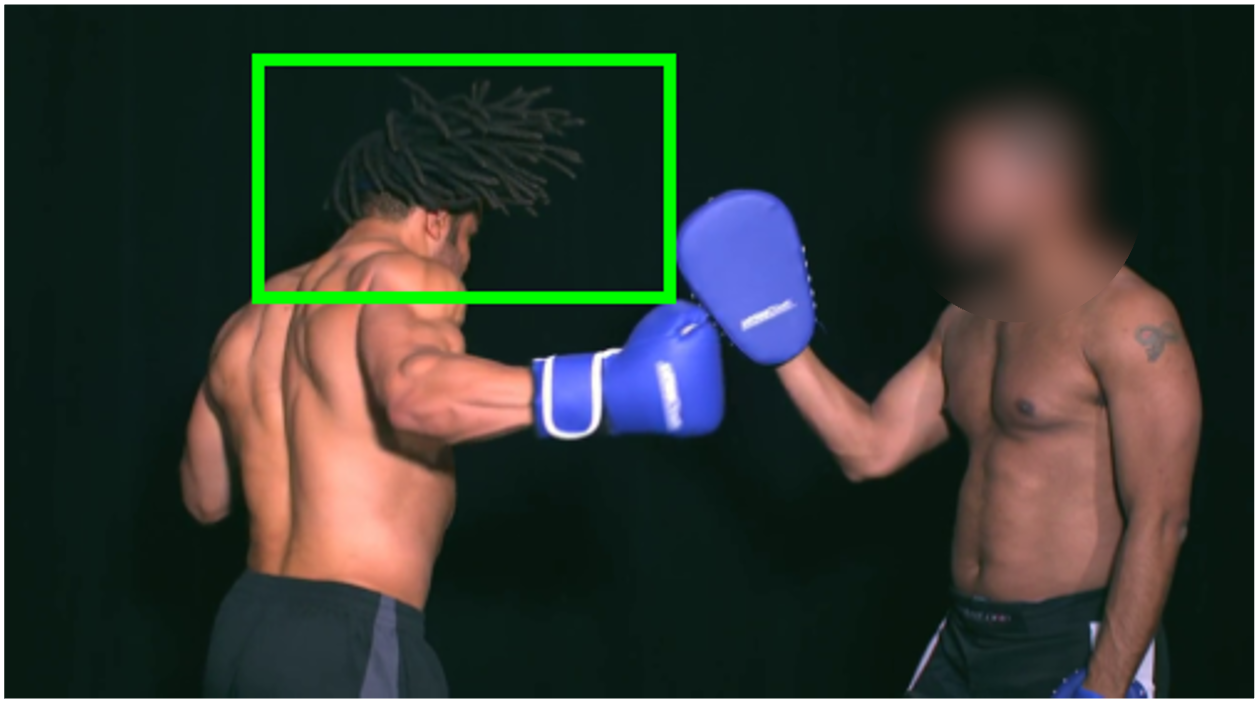} &
    \includegraphics[width=0.198\linewidth]{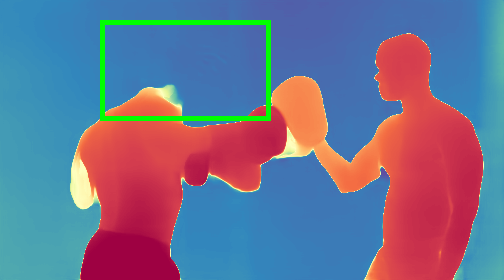} &
    \includegraphics[width=0.198\linewidth]{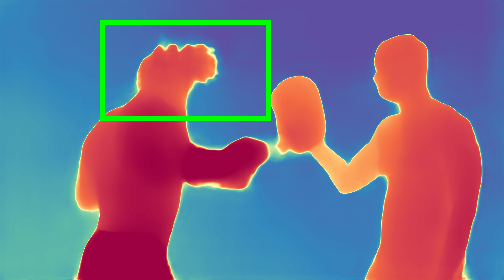} &
    \includegraphics[width=0.198\linewidth]{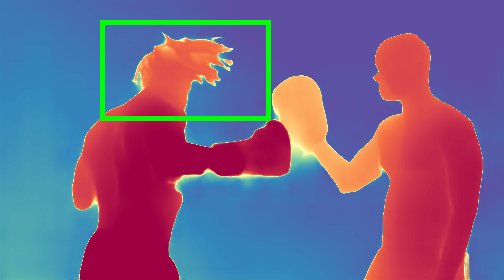} &
    \includegraphics[width=0.198\linewidth]{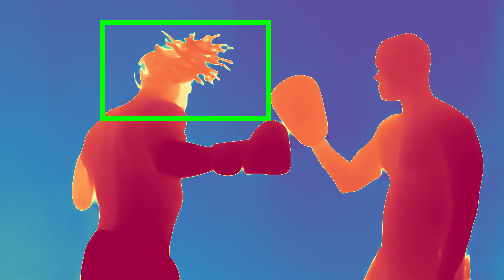} \\
    \includegraphics[width=0.198\linewidth]{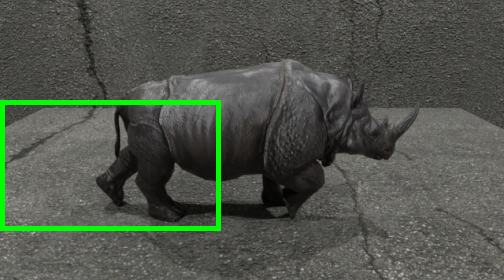} &
    \includegraphics[width=0.198\linewidth]{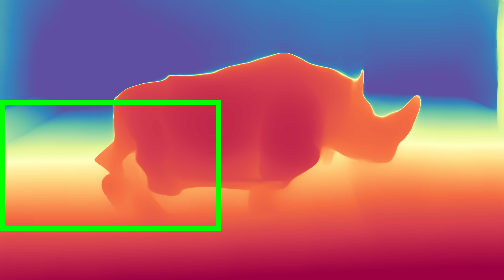} &
    \includegraphics[width=0.198\linewidth]{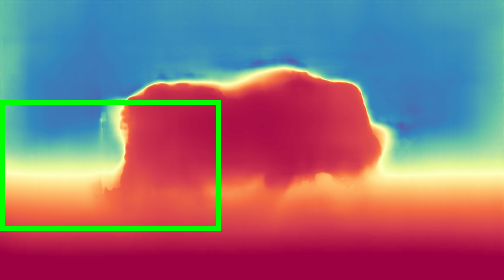} &
    \includegraphics[width=0.198\linewidth]{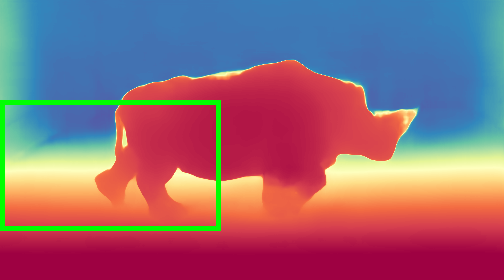} &
    \includegraphics[width=0.198\linewidth]{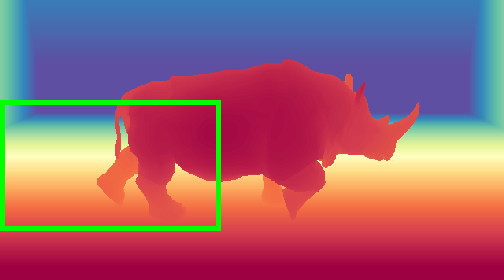} \\
  \end{tabular}
  \caption{Qualitative depth-interpolation comparison with \textit{VDM-EVFI} and \textit{Ours w/o events} at t = 0.5 on DSEC (top), SportsSloMo
  (middle), and Sim4D (bottom).
  Green boxes mark the region of interest in each row.}
  \label{fig:depth_comparison_supp}
\end{figure*}

In~\cref{fig:depth_comparison_supp}, VDM-EVFI exhibits the same limitation as TimeLens.
Because its pipeline is two-stage, first RGB interpolation and then geometry estimation, errors accumulate across the two stages.
As a result, the pillar disappears in the DSEC~\cite{gehrig2021dsec} scene and the player's head is lost in the SportsSloMo~\cite{chen2023sportsslomo} scene.
Interestingly, on Sim4D~\cite{matsuki20254dtam}, VDM-EVFI yields noticeably sharper and cleaner reconstructions than TimeLens.
Being a diffusion-based model, however, it tends to hallucinate fine details: the rhino's nose and legs are sharp but inaccurate when compared to GT.

The results of \textit{Ours w/o events} highlight the importance of the event input and show that our full model actively exploits it.
When the model relies only on the two sparse RGB keyframes to perform feature interpolation, it misplaces the pillar in DSEC and the boxer's fist in SportsSloMo.
Our full model, which uses temporal cues from events, accurately locates objects at intermediate timestamps.
Naturally, our method also yields more accurate boundary reconstruction, as visible in the Sim4D result.

\section{Additional Qualitative Robustness Analysis}
\label{sec:supp_robustness}

\begin{figure*}[t]
  \centering
  \setlength{\tabcolsep}{0.5pt}
  \small
  \newcommand{\rowlabel}[1]{%
    \adjustbox{valign=c}{\rotatebox{90}{#1}}}
  \newcommand{\fig}[1]{%
    \includegraphics[height=2.0cm,valign=c]{#1}}
  \begin{tabular}{cccc}
    & Sharp & Blur & Dark \\
    \rowlabel{\shortstack{TimeLens \\ RGB}} &
    \fig{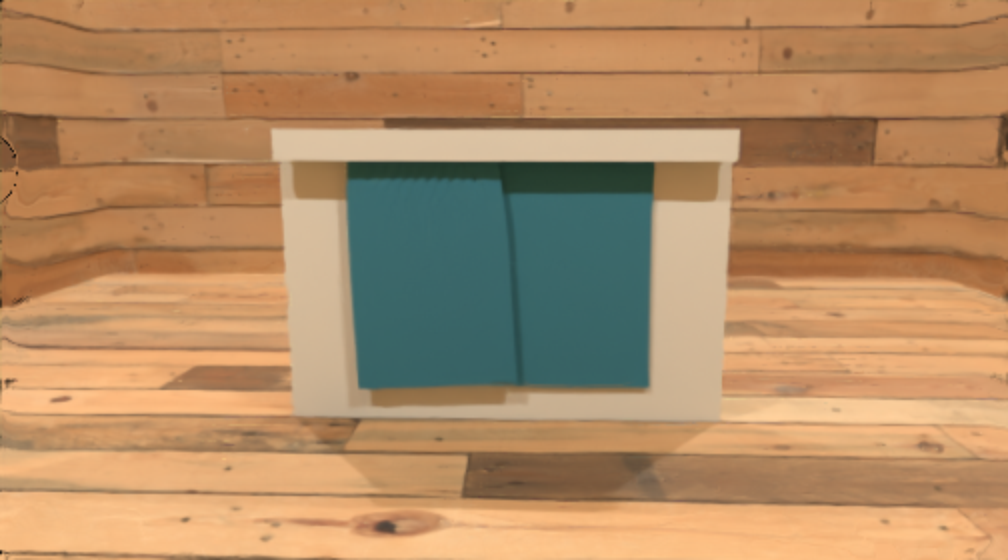} &
    \fig{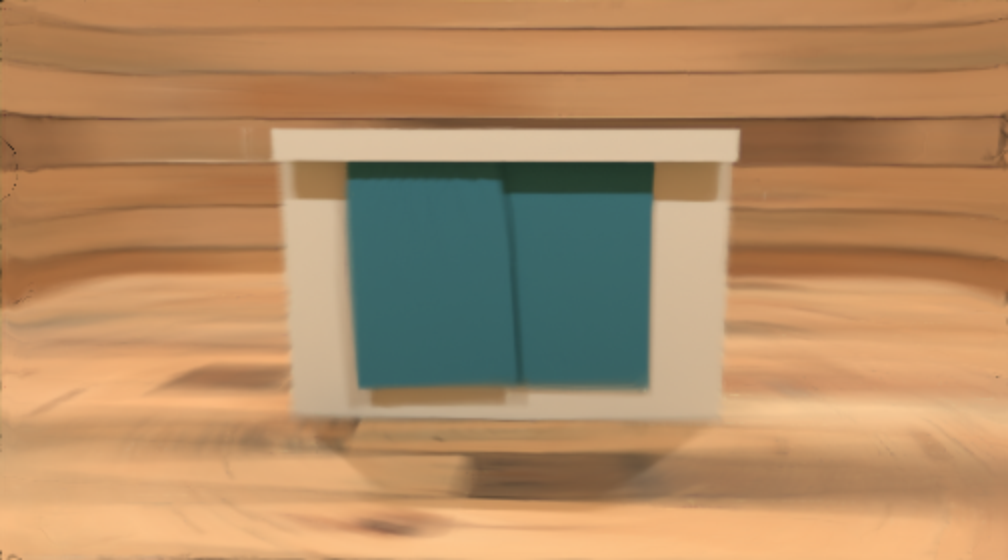} &
    \fig{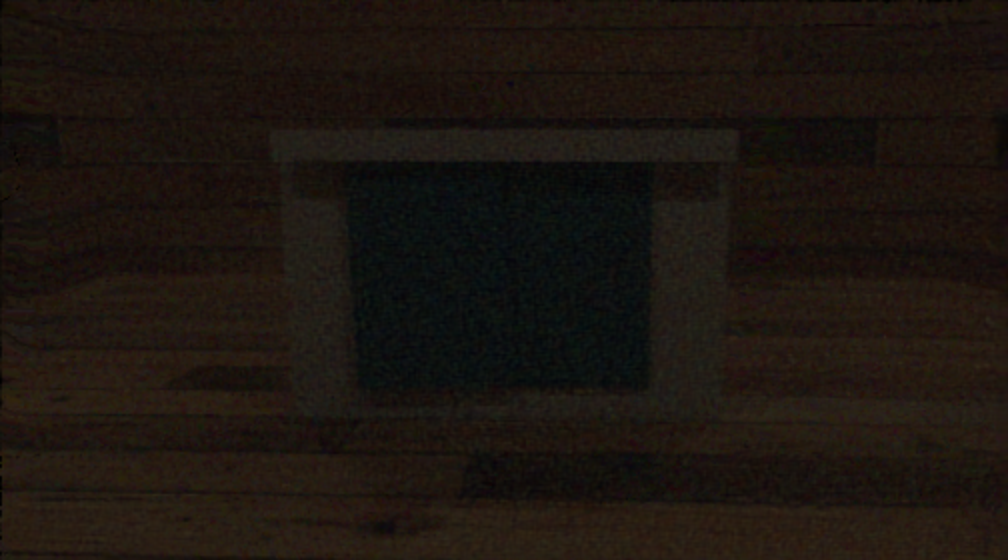} \\
    \rowlabel{\shortstack{TimeLens \\ Pointcloud}} &
    \fig{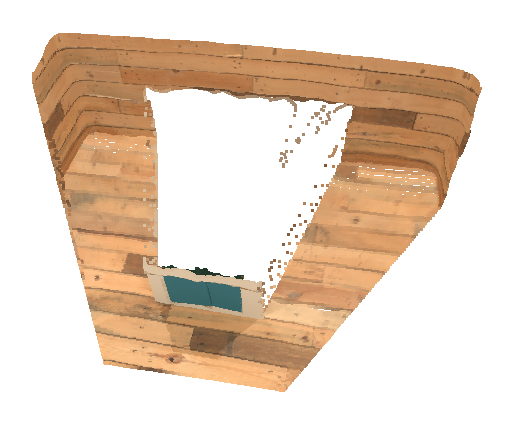} &
    \fig{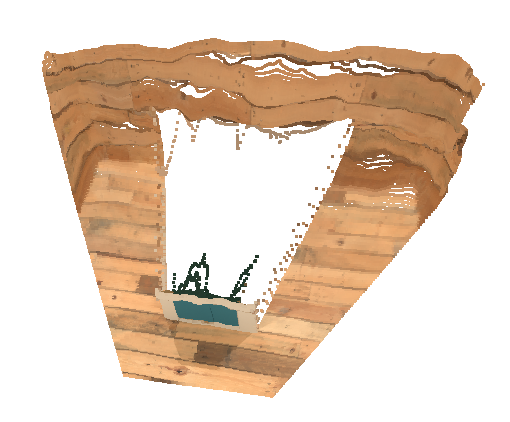} &
    \fig{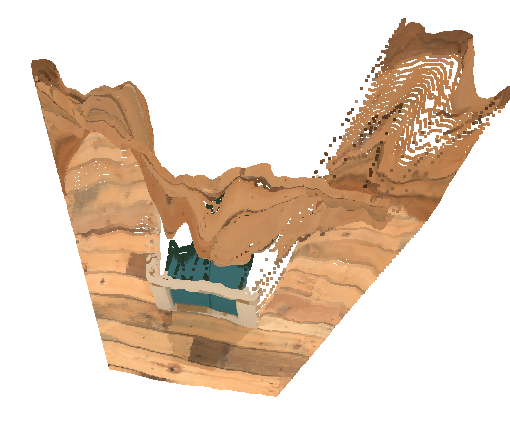} \\
    \rowlabel{\shortstack{VDM-EVFI \\ RGB}} &
    \fig{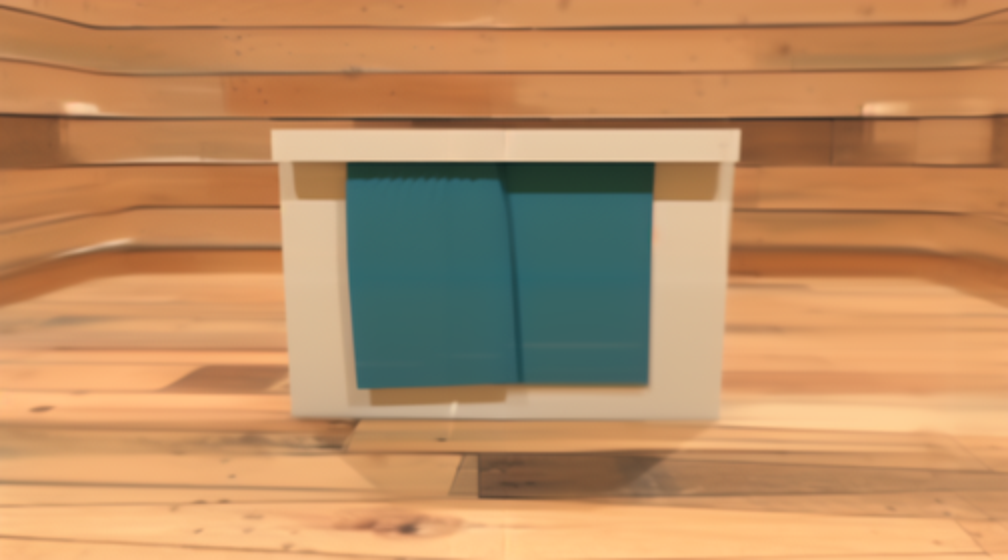} &
    \fig{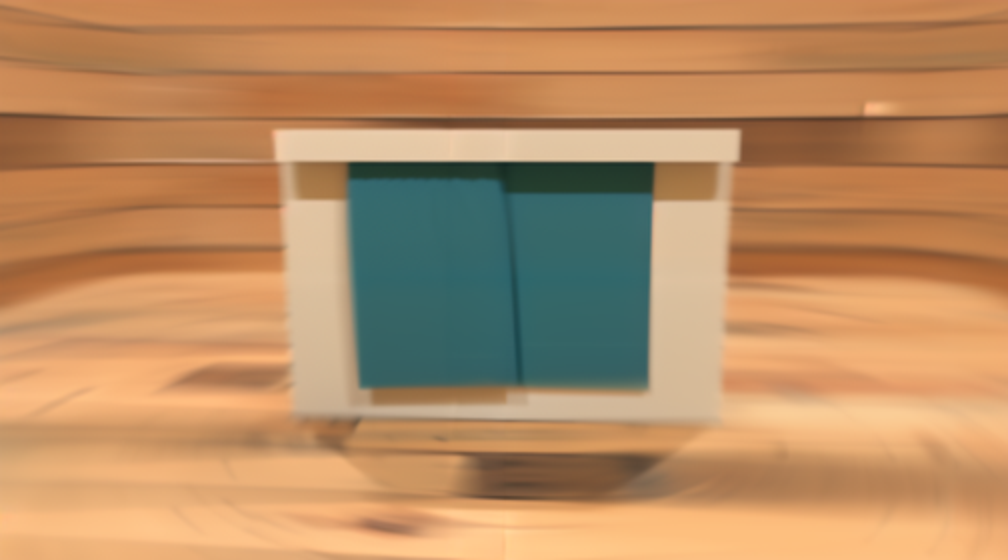} &
    \fig{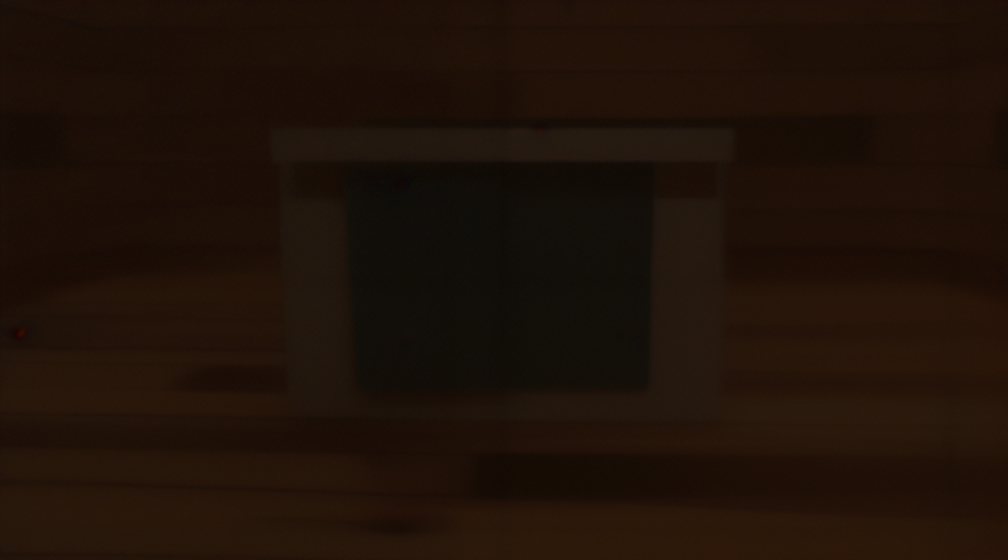} \\
    \rowlabel{\shortstack{VDM-EVFI \\ Pointcloud}} &
    \fig{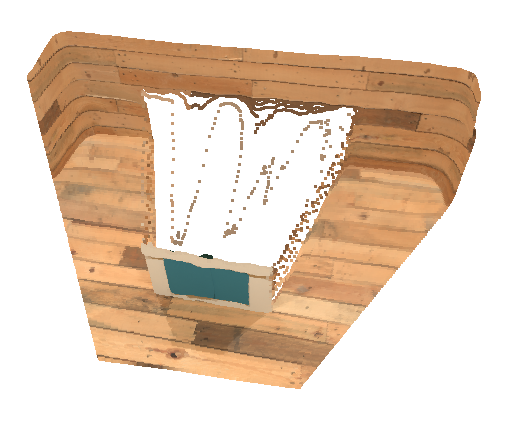} &
    \fig{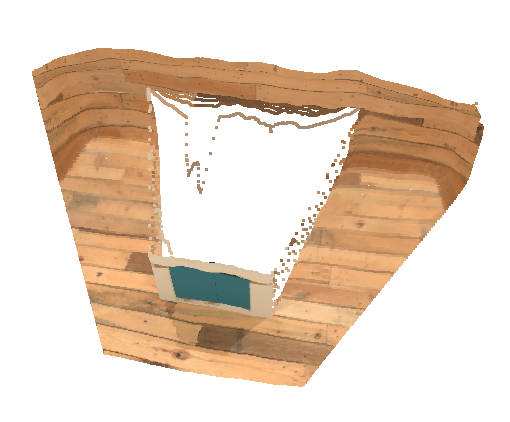} &
    \fig{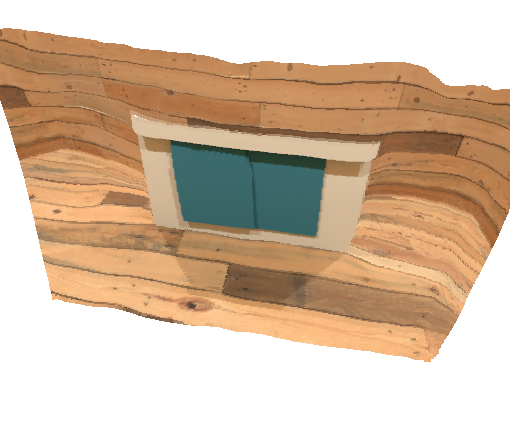} \\
    \rowlabel{\shortstack{Ours \\ RGB}} &
    \fig{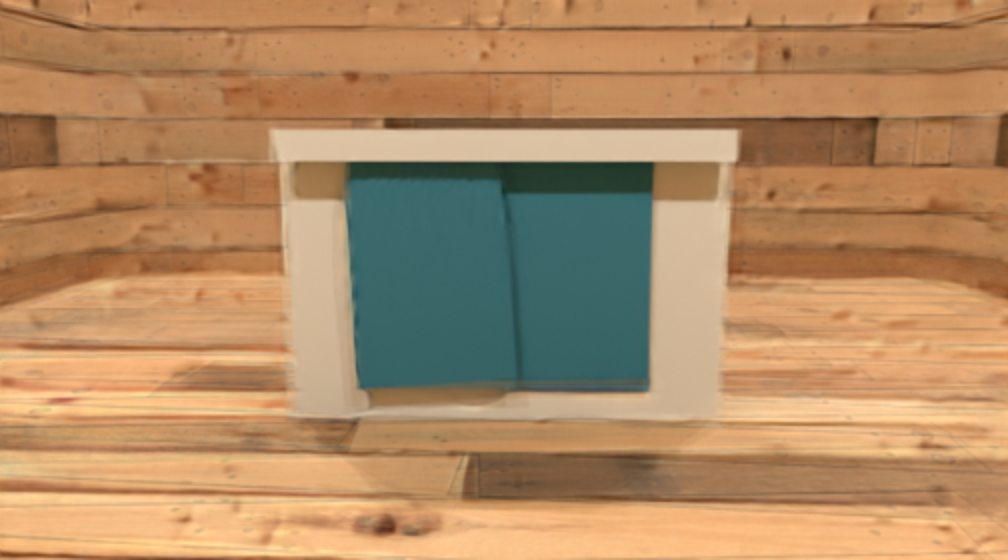} &
    \fig{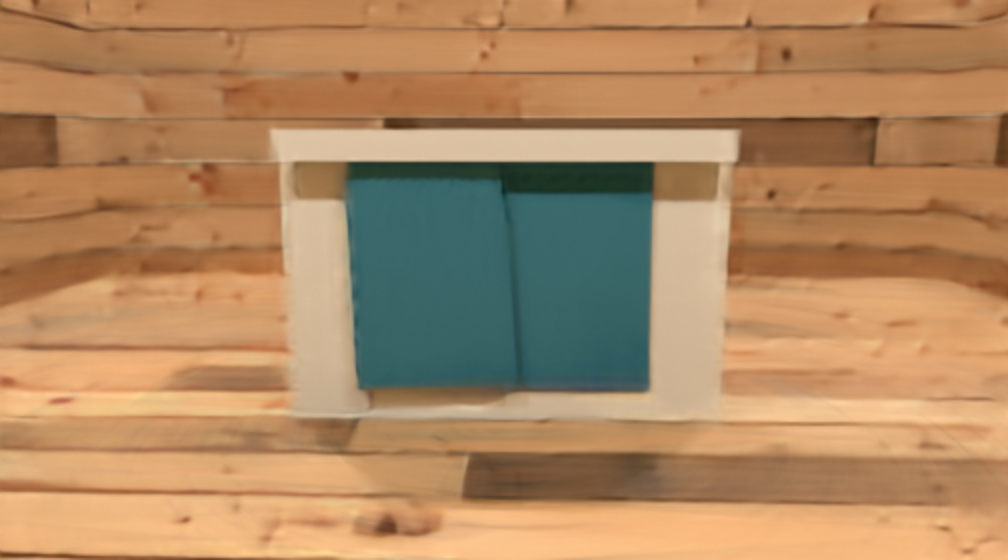} &
    \fig{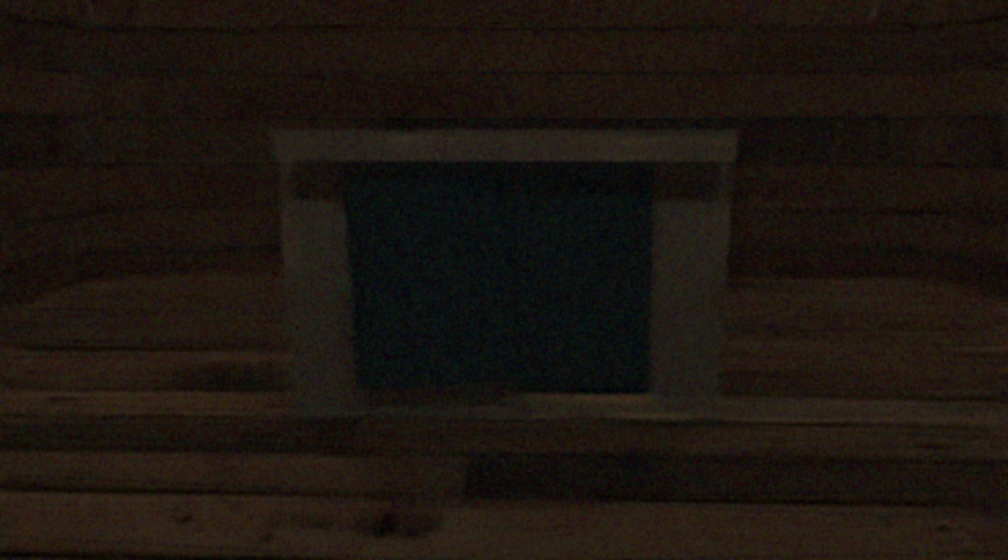} \\
    \rowlabel{\shortstack{Ours \\ Pointcloud}} &
    \fig{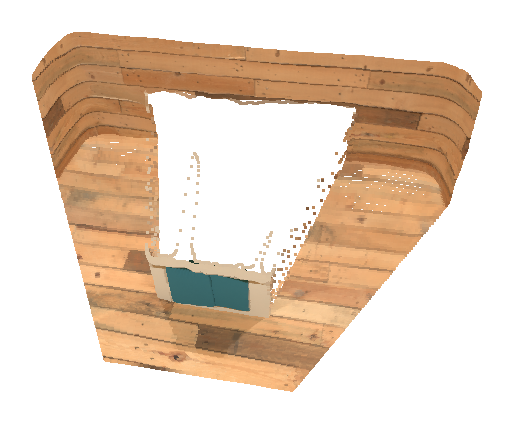} &
    \fig{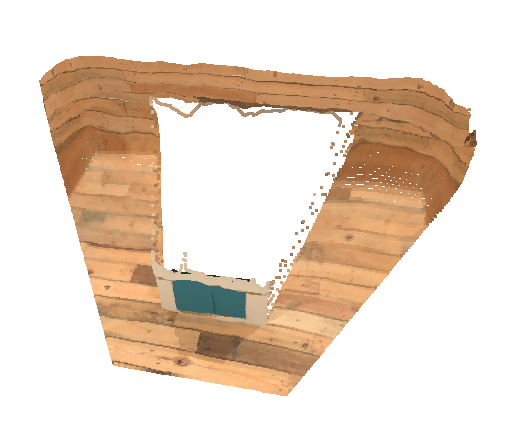} &
    \fig{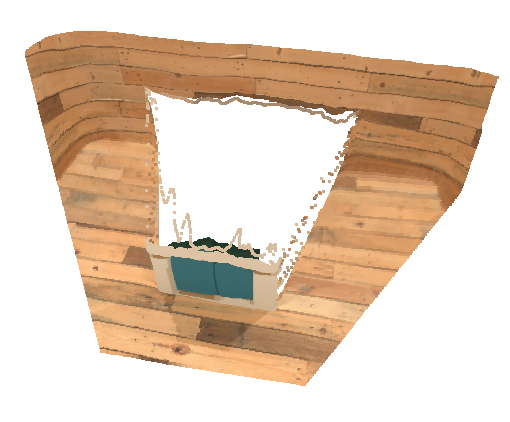} \\
  \end{tabular}
  \caption{Robustness analysis on the Sim4D dataset showing each method's
    interpolated RGB frame at $t = 0.5$ paired with its corresponding
    reconstructed pointcloud.}
  \label{fig:robustness_supp}
\end{figure*}

In~\cref{fig:robustness_supp}, we further analyze color and depth interpolation quality under corrupted inputs.
We compare our method against two event-based RGB interpolation baselines.
For each, we show the interpolated color frame at $t = 0.5$ paired with the corresponding reconstructed point cloud under sharp, motion-blur, and low-light inputs.
Under motion-blur, our method successfully reconstructs sharp intermediate frames, whereas both TimeLens and VDM-EVFI produce blurred interpolations that lead to noise in the point cloud reconstruction.
Under low-light (dark) inputs, both baselines fail to separate the foreground from the background.
When the RGB input is heavily corrupted, interpolating the RGB first and then estimating geometry suffers more from error accumulation.
Although our training data does not contain low-light scenes, our method generalizes naturally to this regime by actively using event information.

\section{Interval-wise Quantitative Analysis on Sim4D}
\label{sec:supp_interval}
\begin{table*}[t]
  \centering
  \small
  \setlength{\tabcolsep}{4pt}
  \begin{tabular}{c l c c c c c c c}
    \toprule
    \multirow{2}{*}{Interval} & \multirow{2}{*}{Method} & \multicolumn{4}{c}{Depth} & \multicolumn{3}{c}{Pose} \\
    \cmidrule(lr){3-6} \cmidrule(lr){7-9}
    & & Abs Rel $\downarrow$ & $\delta < 1.05$ $\uparrow$ & $\delta < 1.10$ $\uparrow$ & $\delta < 1.25$ $\uparrow$ & ATE $\downarrow$ & RTE $\downarrow$ & RRE $\downarrow$ \\
    \midrule
    \multirow{5}{*}{2}
      & TimeLens   & 0.0135          & 0.9704          & \textbf{0.9899} & \textbf{0.9958} & 0.0047          & 0.0593          & 1.9347          \\
      & VDM-EVFI   & 0.0203          & 0.9273          & 0.9782          & 0.9931          & 0.0053          & 0.0588          & 1.8823          \\
      & Interp3R   & 0.0149          & 0.9677          & 0.9829          & 0.9920           & 0.0191          & \textbf{0.0316} & \textbf{1.3329} \\
      & w/o events & 0.0157          & 0.9647          & 0.9838          & 0.9924          & 0.0024          & 0.0576          & 2.0143          \\
      & Ours       & \textbf{0.0121} & \textbf{0.9770}  & 0.9896          & 0.9957          & \textbf{0.0022} & 0.0585          & 1.9381          \\
    \midrule
    \multirow{5}{*}{4}
      & TimeLens   & 0.0140           & 0.9713          & \textbf{0.9886} & \textbf{0.9952} & 0.0062          & 0.0569          & 1.8774          \\
      & VDM-EVFI   & 0.0216          & 0.9229          & 0.9777          & 0.9921          & 0.0118          & 0.0548          & 1.8556          \\
      & Interp3R   & 0.0170           & 0.9609          & 0.9785          & 0.9893          & 0.0245          & \textbf{0.0323} & \textbf{1.3549} \\
      & w/o events & 0.0177          & 0.9593          & 0.9796          & 0.9902          & \textbf{0.0057} & 0.0544          & 1.8653          \\
      & Ours       & \textbf{0.0123} & \textbf{0.9766} & 0.9882          & 0.9951          & 0.0059          & 0.0551          & 1.8174          \\
    \midrule
    \multirow{5}{*}{8}
      & TimeLens   & 0.0194          & 0.9405          & 0.9677          & 0.9905          & \textbf{0.0095} & 0.0587          & 1.9411          \\
      & VDM-EVFI   & 0.0349          & 0.8130           & 0.9332          & 0.9843          & 0.0207          & 0.0560           & 1.7451          \\
      & Interp3R   & 0.0224          & 0.9322          & 0.9699          & 0.9854          & 0.0318          & \textbf{0.0369} & \textbf{1.4909} \\
      & w/o events & 0.0240           & 0.9328          & 0.9704          & 0.9847          & 0.0166          & 0.0527          & 1.7820           \\
      & Ours       & \textbf{0.0150}  & \textbf{0.9650}  & \textbf{0.9850}  & \textbf{0.9932} & 0.0168          & 0.0537          & 1.7561          \\
    \midrule
    \multirow{5}{*}{16}
      & TimeLens   & 0.0397          & 0.8306          & 0.9193          & 0.9615          & \textbf{0.0184} & 0.0627          & 1.8143          \\
      & VDM-EVFI   & 0.0565          & 0.6697          & 0.8440           & 0.9609          & 0.0443          & 0.0586          & 1.6630           \\
      & Interp3R   & 0.0319          & 0.8683          & 0.9551          & 0.9806          & 0.0495          & \textbf{0.0422} & 1.5326          \\
      & w/o events & 0.0398          & 0.8161          & 0.9395          & 0.9747          & 0.0588          & 0.0437          & 1.4361          \\
      & Ours       & \textbf{0.0247} & \textbf{0.8988} & \textbf{0.9674} & \textbf{0.9895} & 0.0583          & 0.0444          & \textbf{1.4254} \\
    \midrule
    \multirow{5}{*}{Average}
      & TimeLens   & 0.0216          & 0.9282          & 0.9664          & 0.9857          & \textbf{0.0097} & 0.0594          & 1.8919          \\
      & VDM-EVFI   & 0.0333          & 0.8332          & 0.9333          & 0.9826          & 0.0205          & 0.0570           & 1.7865          \\
      & Interp3R   & 0.0215          & 0.9323          & 0.9716          & 0.9868          & 0.0312          & \textbf{0.0358} & \textbf{1.4278} \\
      & w/o events & 0.0243          & 0.9182          & 0.9683          & 0.9855          & 0.0209          & 0.0521          & 1.7744          \\
      & Ours       & \textbf{0.0160}  & \textbf{0.9543} & \textbf{0.9826} & \textbf{0.9934} & 0.0208          & 0.0529          & 1.7343          \\
    \bottomrule
  \end{tabular}
  \caption{Quantitative comparison on \textbf{Sim4D-Sharp}.
    Depth and pose interpolation under sharp imaging conditions, across query intervals of 2, 4, 8, and 16 frames and their average.
    The best result in each column is in bold.}
  \label{tab:supp_sharp}
\end{table*}

\begin{table*}[t]
  \centering
  \small
  \setlength{\tabcolsep}{4pt}
  \begin{tabular}{c l c c c c c c c}
    \toprule
    \multirow{2}{*}{Interval} & \multirow{2}{*}{Method} & \multicolumn{4}{c}{Depth} & \multicolumn{3}{c}{Pose} \\
    \cmidrule(lr){3-6} \cmidrule(lr){7-9}
    & & Abs Rel $\downarrow$ & $\delta < 1.05$ $\uparrow$ & $\delta < 1.10$ $\uparrow$ & $\delta < 1.25$ $\uparrow$ & ATE $\downarrow$ & RTE $\downarrow$ & RRE $\downarrow$ \\
    \midrule
    \multirow{5}{*}{2}
      & TimeLens   & 0.0533          & 0.7232          & 0.8580           & 0.9468          & 0.0103          & 0.0518          & 1.2194          \\
      & VDM-EVFI   & 0.0533          & 0.7195          & 0.8657          & 0.9533          & 0.0081          & 0.0524          & \textbf{1.1683} \\
      & Interp3R   & 0.0494          & 0.7368          & 0.8794          & 0.9533          & 0.0183          & \textbf{0.0311} & 1.2963          \\
      & w/o events & 0.0519          & 0.7151          & 0.8761          & 0.9571          & 0.0083          & 0.0509          & 1.6481          \\
      & Ours       & \textbf{0.0431} & \textbf{0.7618} & \textbf{0.8994} & \textbf{0.9709} & \textbf{0.0068} & 0.0522          & 1.5379          \\
    \midrule
    \multirow{5}{*}{4}
      & TimeLens   & 0.0556          & 0.7073          & 0.8468          & 0.9432          & 0.0148          & 0.0538          & 1.2670           \\
      & VDM-EVFI   & 0.0549          & 0.7023          & 0.8556          & 0.9530           & 0.0141          & 0.0536          & \textbf{1.2048} \\
      & Interp3R   & 0.0435          & 0.7916          & 0.9001          & 0.9549          & 0.0253          & \textbf{0.0325} & 1.3277          \\
      & w/o events & 0.0485          & 0.7464          & 0.9006          & 0.9576          & 0.0115          & 0.0519          & 1.5502          \\
      & Ours       & \textbf{0.0364} & \textbf{0.8216} & \textbf{0.9263} & \textbf{0.9725} & \textbf{0.0101} & 0.0532          & 1.5177          \\
    \midrule
    \multirow{5}{*}{8}
      & TimeLens   & 0.0551          & 0.7458          & 0.8714          & 0.9363          & \textbf{0.0175} & 0.0577          & 1.4619          \\
      & VDM-EVFI   & 0.0557          & 0.6793          & 0.8366          & 0.9599          & 0.0204          & 0.0569          & \textbf{1.3456} \\
      & Interp3R   & 0.0447          & 0.7605          & 0.8947          & 0.9599          & 0.0363          & \textbf{0.0383} & 1.3992          \\
      & w/o events & 0.0531          & 0.7064          & 0.8813          & 0.9577          & 0.0183          & 0.0537          & 1.5964          \\
      & Ours       & \textbf{0.0366} & \textbf{0.8074} & \textbf{0.9254} & \textbf{0.9777} & \textbf{0.0175} & 0.0541          & 1.5659          \\
    \midrule
    \multirow{5}{*}{16}
      & TimeLens   & 0.0673          & 0.6161          & 0.7948          & 0.9371          & \textbf{0.0281} & 0.0609          & 1.6073          \\
      & VDM-EVFI   & 0.0701          & 0.5907          & 0.7779          & 0.9391          & 0.0332          & 0.0619          & 1.5493          \\
      & Interp3R   & 0.0590           & 0.6629          & 0.8298          & 0.9546          & 0.0520           & 0.0442          & 1.4454          \\
      & w/o events & 0.0660           & 0.6243          & 0.8279          & 0.9491          & 0.0578          & \textbf{0.0438} & \textbf{1.4163} \\
      & Ours       & \textbf{0.0472} & \textbf{0.7079} & \textbf{0.8747} & \textbf{0.9733} & 0.0579          & 0.0454          & 1.4426          \\
    \midrule
    \multirow{5}{*}{Average}
      & TimeLens   & 0.0579          & 0.6981          & 0.8428          & 0.9408          & \textbf{0.0177} & 0.0560           & 1.3889          \\
      & VDM-EVFI   & 0.0585          & 0.6729          & 0.8339          & 0.9513          & 0.0190           & 0.0562          & \textbf{1.3170}  \\
      & Interp3R   & 0.0492          & 0.7379          & 0.8760           & 0.9557          & 0.0330           & \textbf{0.0365} & 1.3672          \\
      & w/o events & 0.0549          & 0.6980           & 0.8715          & 0.9554          & 0.0240           & 0.0501          & 1.5527          \\
      & Ours       & \textbf{0.0408} & \textbf{0.7747} & \textbf{0.9064} & \textbf{0.9736} & 0.0231          & 0.0512          & 1.5160           \\
    \bottomrule
  \end{tabular}
  \caption{Quantitative comparison on \textbf{Sim4D-Blur}.
    Depth and pose interpolation under motion-blur imaging conditions, across query intervals of 2, 4, 8, and 16 frames and their average.
    The best result in each column is in bold.}
  \label{tab:supp_blur}
\end{table*}

\begin{table*}[t]
  \centering
  \small
  \setlength{\tabcolsep}{4pt}
  \begin{tabular}{c l c c c c c c c}
    \toprule
    \multirow{2}{*}{Interval} & \multirow{2}{*}{Method} & \multicolumn{4}{c}{Depth} & \multicolumn{3}{c}{Pose} \\
    \cmidrule(lr){3-6} \cmidrule(lr){7-9}
    & & Abs Rel $\downarrow$ & $\delta < 1.05$ $\uparrow$ & $\delta < 1.10$ $\uparrow$ & $\delta < 1.25$ $\uparrow$ & ATE $\downarrow$ & RTE $\downarrow$ & RRE $\downarrow$ \\
    \midrule
    \multirow{5}{*}{2}
      & TimeLens   & 0.0722          & 0.5913          & 0.7890           & 0.9314          & 0.0082          & 0.0549          & 1.6552          \\
      & VDM-EVFI   & 0.0829          & 0.5477          & 0.7466          & 0.9148          & 0.0101          & 0.0517          & 1.4927          \\
      & Interp3R   & 0.0579          & 0.6360           & 0.8304          & 0.9644          & 0.0187          & \textbf{0.0317} & \textbf{1.2799} \\
      & w/o events & 0.0521          & 0.6764          & 0.8741          & 0.9686          & 0.0049          & 0.0558          & 1.8151          \\
      & Ours       & \textbf{0.0481} & \textbf{0.7140}  & \textbf{0.8874} & \textbf{0.9725} & \textbf{0.0037} & 0.0562          & 1.7591          \\
    \midrule
    \multirow{5}{*}{4}
      & TimeLens   & 0.0772          & 0.5733          & 0.7655          & 0.9189          & 0.0149          & 0.0540           & 1.6005          \\
      & VDM-EVFI   & 0.1115          & 0.4416          & 0.6348          & 0.8521          & 0.0208          & 0.0521          & 1.4358          \\
      & Interp3R   & 0.0575          & 0.6472          & 0.8345          & 0.9639          & 0.0257          & \textbf{0.0324} & \textbf{1.4223} \\
      & w/o events & 0.0505          & 0.7025          & 0.8798          & 0.9675          & 0.0083          & 0.0543          & 1.7207          \\
      & Ours       & \textbf{0.0461} & \textbf{0.7315} & \textbf{0.8930}  & \textbf{0.9740}  & \textbf{0.0072} & 0.0541          & 1.6896          \\
    \midrule
    \multirow{5}{*}{8}
      & TimeLens   & 0.0948          & 0.5413          & 0.7234          & 0.8602          & 0.0271          & 0.0567          & 1.5838          \\
      & VDM-EVFI   & 0.1979          & 0.2897          & 0.4147          & 0.6372          & 0.0415          & 0.0542          & 1.6513          \\
      & Interp3R   & 0.0658          & 0.6098          & 0.7985          & 0.9492          & 0.0328          & \textbf{0.0366} & \textbf{1.5779} \\
      & w/o events & 0.0527          & 0.7086          & 0.8648          & 0.9637          & \textbf{0.0183} & 0.0521          & 1.6512          \\
      & Ours       & \textbf{0.0467} & \textbf{0.7360}  & \textbf{0.8832} & \textbf{0.9723} & 0.0184          & 0.0530           & 1.6180           \\
    \midrule
    \multirow{5}{*}{16}
      & TimeLens   & 0.1163          & 0.4501          & 0.6366          & 0.8191          & \textbf{0.0517} & 0.0611          & 1.4717          \\
      & VDM-EVFI   & 0.2407          & 0.1869          & 0.3099          & 0.5573          & 0.0647          & 0.0640           & 1.9151          \\
      & Interp3R   & 0.0843          & 0.5232          & 0.7311          & 0.9202          & 0.0518          & \textbf{0.0417} & 1.6187          \\
      & w/o events & 0.0679          & 0.6032          & 0.8075          & 0.9474          & 0.0613          & 0.0438          & \textbf{1.3578} \\
      & Ours       & \textbf{0.0560}  & \textbf{0.6939} & \textbf{0.8488} & \textbf{0.9616} & 0.0612          & 0.0448          & 1.3704          \\
    \midrule
    \multirow{5}{*}{Average}
      & TimeLens   & 0.0901          & 0.5390           & 0.7286          & 0.8824          & 0.0255          & 0.0567          & 1.5778          \\
      & VDM-EVFI   & 0.1582          & 0.3665          & 0.5265          & 0.7403          & 0.0343          & 0.0555          & 1.6237          \\
      & Interp3R   & 0.0664          & 0.6040           & 0.7986          & 0.9494          & 0.0323          & \textbf{0.0356} & \textbf{1.4747} \\
      & w/o events & 0.0558          & 0.6727          & 0.8565          & 0.9618          & 0.0232          & 0.0515          & 1.6362          \\
      & Ours       & \textbf{0.0492} & \textbf{0.7189} & \textbf{0.8781} & \textbf{0.9701} & \textbf{0.0226} & 0.0520           & 1.6093          \\
    \bottomrule
  \end{tabular}
  \caption{Quantitative comparison on \textbf{Sim4D-Dark}.
    Depth and pose interpolation under low-light imaging conditions, across query intervals of 2, 4, 8, and 16 frames and their average.
    The best result in each column is in bold.}
  \label{tab:supp_dark}
\end{table*}

In the main paper, we conduct an interval-wise quantitative comparison on the DSEC dataset (\cref{tab:dsec}), but report only average values for Sim4D (\cref{tab:sim4d}).
In this section, we provide an interval-wise quantitative analysis on the Sim4D dataset for all three input types: sharp, motion-blur, and low-light.
For Sim4D-Sharp (\cref{tab:supp_sharp}), our method and the baselines achieve comparable depth errors at small intervals, but our method demonstrates superior performance than the baselines as the interval grows.
For Sim4D-Blur (\cref{tab:supp_blur}) and Sim4D-Dark (\cref{tab:supp_dark}), our method already outperforms the baselines by a large margin even at small intervals.
On these corrupted inputs, our method achieves the best performance across all depth metrics at every interval.
In terms of global pose error (ATE), our method is worse than the image-space baselines on sharp images.
We attribute this to the simple scene structure of Sim4D.
However, this gap narrows on Sim4D-Blur and reverses on Sim4D-Dark, where our method performs best.

\end{document}